\documentclass[acmsmall,authorversion=true]{acmart}
\AtBeginDocument{%
  \providecommand\BibTeX{{%
    \normalfont B\kern-0.5em{\scshape i\kern-0.25em b}\kern-0.8em\TeX}}}

\acmDOI{10.1145/3450445}
\acmMonth{5}
\acmYear{2021}
\acmJournal{TWEB}
\acmVolume{15}
\acmNumber{2}
\acmArticle{9}

\if{0}
\setcopyright{acmcopyright}
\copyrightyear{2019}
\acmYear{2019}
\acmDOI{xx.xx/xxxx.xxxx}

\acmJournal{JACM}
\acmVolume{37}
\acmNumber{4}
\acmArticle{111}
\acmMonth{8}
\fi

\usepackage{tcolorbox}
\usepackage{enumerate}
\usepackage{xcolor}
\usepackage[normalem]{ulem}

\begin{document}

\title{An Integrated Approach for Improving Brand Consistency of Web Content: Modeling, Analysis and Recommendation}


 \author{Soumyadeep Roy}
\affiliation{
\institution{IIT Kharagpur, India}}
\email{soumyadeep.roy9@iitkgp.ac.in}

\author{Shamik Sural}
\affiliation{
\institution{IIT Kharagpur, India}}
\email{shamik@cse.iitkgp.ac.in}

\author{Niyati Chhaya}
\affiliation{\institution{Adobe Research, India}}
\email{nchhaya@adobe.com}

\author{Anandhavelu Natarajan}
\affiliation{\institution{Adobe Research, India}}
\email{anandvn@adobe.com}

 \author{Niloy Ganguly}
\affiliation{
\institution{IIT Kharagpur, India}}
\email{niloy@cse.iitkgp.ac.in}

\renewcommand{\shortauthors}{S. Roy et al.}

\begin{abstract}
  \noindent A consumer-dependent (business-to-consumer) organization tends to present itself as possessing a set of human qualities, which is termed as the \textit{brand personality} of the company. The perception is impressed upon the consumer through the content, be it in the form of advertisement, blogs or magazines, produced by the organization. A consistent brand will generate trust and retain customers over time as they develop an affinity towards regularity and common patterns.
  However, maintaining a consistent messaging tone for a brand has become more challenging with the virtual explosion in the amount of content which needs to be authored and pushed to the Internet to maintain an edge in the era of digital marketing. To understand the depth of the problem,  
 we collect around $300 K$ web page content from around $650$ companies. We develop trait-specific classification models by considering the linguistic features of the content. 
   The classifier automatically identifies the web articles which are not consistent with the mission and vision of a company and further helps us to discover the conditions under which the consistency cannot be maintained. To address the brand inconsistency issue, we then develop a sentence ranking system that outputs the top three sentences that need to be changed for making a web article more consistent with the company's brand personality. 
\end{abstract}

\ccsdesc[300]{Computing methodologies~Ranking}
\ccsdesc[100]{Computing methodologies~Supervised learning by classification}
\ccsdesc[300]{Information systems~Web mining}

\keywords{brand personality, online reputation management, sentence ranking, text classification}

\maketitle

 \section{Introduction}\label{sec:introduction}
In the era of digital marketing, it is becoming increasingly challenging for organizations to position their brand uniquely and differentiate from competitive product lines. In order to achieve that, organizations target a set of humanlike qualities (or a personality) across all their marketing campaigns. For example, Red Bull portrays itself as courageous and uninhibited, while Nike represents itself as being athletic. These brand personality traits have been formalized by Aaker~\cite{aaker1997} who outlined the five dimensions along with its characteristics (see  Table~\ref{table:brandtraits})~\footnote{It is one of the three constituents that make up the brand image of an organization, along with {\em product attributes} and {\em consumer benefits}.}.
An important way of portraying the personality is by carefully drafting the posts that are regularly updated in the company's web portal. Note that large companies have continuous and voluminous updates.
In order to ensure that the contents added are consistent with the brand image of an organization,
the large scale content authoring needs to be monitored. This is presently checked manually with the brand style guide of a company. 
Currently, companies~\footnote{We use the term `company' and `organization' interchangeably in the paper.} issue a brand style guide which the brand managers and content writers follow. It has multiple components - colour, typography, logo, the position of headers and content, personality and tone and voice (like in Mozilla's brand style guide~\footnote{https://mozilla.design}).
In this work, we specifically deal with the {\em tone and voice} aspect of a brand style guide, which affects the textual content of any such communication. In terms of consumer psychology, it is termed as {\em brand personality}. 

\begin{table}[!ht]
\centering
\begin{tabular}{ c|p{10cm}}
\hline
Trait&Characteristics\\
 \hline
 sincerity & honest, friendly, sincere or down-to-earth.\\
 excitement & spirited, imaginative, trendy, and contemporary.\\
 competence & competitive, ambitious, one who focuses on its successes and achievements. \\
 ruggedness & adventurous, outdoorsy, tough or western.\\
 sophistication & glamorous, charming or catering to the financially affluent population.\\

\end{tabular}
\caption{Brand Personality dimensions formulated by Aaker~\cite{aaker1997}}
\label{table:brandtraits}
\end{table}

The web articles are responsible in portraying the brand personality of a company. 
Since consumers look for regularity and common patterns over time within an organization's promotional material, a consistent brand will generate trust and retain customers over a longer period.
To sum up, the brand impression is created by static pages while the dynamic pages play a crucial role in maintaining that impression on the customers and ensuring brand consistency.

We aim to address the issue of maintaining brand personality at the content-creation stage~(performed by brand managers and content creators), contrary to the field of \textit{online reputation monitoring~(ORM)} that deals only with the consumer-generated content. 
Therefore, as our second objective, we develop a helper tool that recommends the top three sentences that should be modified for improving the brand consistency of an article. In simple terms, the helper tool suggests sentence-level changes that will make the web article~(a dynamic post) more aligned to the organization's target brand personality~(created by the company's static posts).

We enlist our contributions below.

\begin{enumerate}[(i)]
\item We select a novel set of linguistic features for the brand personality prediction task. 
To the best of our knowledge, this is the first work that considers the content of official website of Fortune 1000 companies and characterizes brand personality of the companies. We use the \textit{Linguistic Inquiry and Word Count}~(LIWC) score and further propose additional linguistic features such as \textit{Contractions, Collocations, Chains of reference and Flesch's Readability Ease score} for developing our brand personality classification models.
\item We address the problem of unavailability of large amount of high-quality labelled data for the different dimensions of brand personality. Although we create a small human-tagged dataset, it is insufficient to train data-hungry deep learning models or even conduct a large-scale characterization study. We use a classification model to tag the unlabelled articles and consider those data points that are classified with very high confidence. This leads to the formation of a high fidelity~(machine-tagged) set of around 95,536 points (articles), where each article has at least one brand personality trait present with high confidence; we refer to it as MT\textsubscript{high}.
\item We quantify brand consistency as the similarity score computed between the article brand personality and the target brand personality. It is represented in terms of a binary vector (a five-dimensional binary vector where a dimension corresponds to a  brand personality traits), as well as a rank vector~(captures the relative order among the brand personality traits in terms of the degree of presence in the text). We further define three levels of brand consistency~(strongly consistent, consistent, and not consistent) and a temporal formulation of brand consistency, which is more fine-grained, based on the temporal binning strategy of 3 months duration.
\item We perform a large scale characterization study using MT\textsubscript{high} and also study the impact of various company-level factors (see Section~\ref{sec:interbrand}) like Fortune 1000 company rank, presence of brand extensions (multiple product lines), on the overall brand consistency score of a company. We observe that promotion posts irrespective of brand, primarily portray the brand personality - {\em competence}  and favours the brand consistency of companies which portray {\em competence} in most of their posts (primary trait). We find that presence of brand extension-related posts reduces the brand consistency score of a company (see Section~\ref{sec:static-dyn-posts}).
\item We develop a novel sentence-ranking scheme based on the brand consistency, polarity score and centrality aspects, to recommend the top three sentences of a web article that need to be modified in order to improve the post-level brand consistency score. We, therefore, investigate the causes of failure of brand consistency at sentence-level by performing a detailed qualitative study over the posts. Due to the lack of available annotated data for our sentence ranking task, we develop the annotation guidelines for the ground truth creation and experimental setup. We observe that our {\em Multi-Aspect Sentence Ranking} model ({\em MASR-3}) 
achieves the best performance among the baselines except for centrality aspect, in terms of ROUGE-F1, and Precision (at various ranks). We further establish that the centrality aspect is crucial for the sentence ranking task\footnote{The classification approach, development of high fidelity data points and the formulation of brand consistency have been covered in a prior work~\cite{Roy2019BrandPersona}. The current work extends that work in the following manner --- (i) We add a detailed characterization study for inter-company relationships and also study the contributing factors affecting the company-wise brand consistency score. (ii) We formulate our sentence-level ranking task and then design the experimental setup and corresponding annotation guidelines. (iii) We propose a novel sentence ranking scheme~(based on brand consistency, negative polarity and centrality aspects) that recommends the top three sentences of a web article that needs to be modified to improve the brand consistency score of the article. Our model outperforms all competing baselines across a wide range of evaluation metrics; we also perform extensive error analysis.
}.

\end{enumerate}

\noindent \textbf{Organization.} Section \ref{sec:priorart} covers the related works. Section \ref{sec:features} describes the large-scale data collection, corpus analysis, annotation details, followed by the linguistic feature generation step. Section \ref{sec:classifier-model} presents the proposed classification model followed by experimental results. In Section~\ref{brandconsanalysis}, we perform a large scale characterization study of brand consistency. We then study the effect of various company-level factors on the brand consistency score of an organization. We then describe the ground truth creation, and experimental results regarding the sentence ranking tool in Section~\ref{recommendertool} and finally conclude the paper in Section~\ref{sec:conclusion}.

\section{Prior Research}\label{sec:priorart}
We first define the notion of brand personality and brand image in terms of existing `consumer psychology' models and then discuss the current strategies that are used for maintaining it over time. Here, we further highlight the need and current relevance of quantifying brand consistency. We then explore the various computational approaches that are used for brand personality detection. Finally, we discuss prior research studies highlighting the importance of our proposed sentence ranking model, developed to improve the brand consistency of a document, given a target brand personality vector.

\subsection{Brand personality in terms of consumer psychology models}\label{sec:brand-cons-psych-related}
We have earlier defined brand personality as a set of humanlike qualities or a personality that a company wants to maintain in its promotional materials. Digman et al.~\cite{digman1990} formalized this concept of human personality into the Big Five personality traits which are extroversion, neuroticism, agreeableness, conscientiousness and openness. Muller and Chandon~\cite{muller2003impact} study the impact of visiting a company website on the brand image and particularly on the brand personality since companies strategically place verbal or non-verbal cues in the websites to evoke specific emotions from their users. Chen and Rogers~\cite{Chen2006} develop a scale to measure the website personality, using a combination of human personality and brand personality attributes. Douglas et al. ~\cite{douglas2007exploring} proposed a `Website Emotional Features Assessment Model' that is based on website features - site activation, site affection, site confidence, site serenity, site superiority, and site surgency. From the user's perspective, a recent study by Still~\cite{Still2018} found that the visual hierarchy of a webpage depends on the position, colour and text style of each webpage element and not size.

Schmitt et al.~\cite{schmitt2012consumer} proposed a consumer psychology model of brands, which specifically focuses on the unique characteristics of the brands in contrast to the earlier `general information processing'-based models. It primarily identifies that consumers have different levels of psychological engagement with brands due to their individual needs, motives and goal. Brand personality belongs to the middle layer out of the $3$ levels of engagement~(the innermost level is object-centred and functionally-driven, and the outermost layer represents the social interaction and community perspective of a brand), where the brand is seen as personally relevant to the customer. This model also distinguishes among different brand-related issues associated with brand personality. First, the \textit{integrating process} where brand information is combined into an overall brand concept, brand personality, and relationship with the brand. Secondly, the \textit{connecting phase} where the customer forms an attitude towards the brand, becomes personally attached to it, and get connected in a brand community. 

\subsection{Managing brand personality over time and the need for brand consistency}
A strong brand is a powerful driver of sales, profits and shareholder value. However, even the most popular brands may face the danger of getting lost unless they reflect changing customer preferences~\cite{Keller1999}. Keller~\cite{Keller2009} explore the consumer-based brand equity model and explain that understanding consumer's brand knowledge structures is crucial for maintaining brand equity.
Various strategies exist for managing an organization's brand image. They include brand reinforcement, maintaining brand consistency, fine-tuning the support program, importance of timing of product announcements, brand revitalizing, expanding brand awareness, improving brand image, as well as adding new customers and brand extensions~\cite{Keller1999}. Devecik et al.~\cite{brandDivestment2014} study the effect of extreme strategies like \textit{brand divestment} (brand deletion) on the company's stock price. In this paper, we concentrate on studying various aspects responsible for maintaining brand consistency. 

Consumers, in general, tend to collaborate with organizations considering it a living human being; thus, they attempt to replicate their inter-personal and social relationships with organizations too. Consequently, when an organization's activity abuses the relationship standards, customers develop an increasingly negative view of the brand when contrasted with when the brand activities are steady with those relationship norms~\cite{aggarwal2004effects}. InterBrand~\footnote{https://www.interbrand.com/best-brands/} - a world-renowned brand agency, believed consistency to be among the ten qualities of organizations, answerable for sustainable development of brands among different properties like clarity, commitment, governance, responsiveness, authenticity, relevance, differentiation, presence and engagement.

\subsection{Computational approaches to brand personality detection}\label{sec:brand-comput-related}
Aaker \cite{aaker1997} formalizes the notion of brand personality into five dimensions. Xu et al.~\cite{Xu2016} develop a classification model based on user imagery and official announcements on Twitter and employee imagery from Glassdoor, for predicting the perceived brand personality from the textual content. They observe that the consumption behaviour and personality traits depend on the income and needs of the consumer. Liu et al.~\cite{Liu2016, Liu2017} explain the association between brand personality and the previously mentioned factors on social media, with the help of a visual analysis tool. Wu et al.~\cite{brandpersonmobileui} develop a computational model for detecting brand personality from a user interface of a mobile application based on both element and page-level user interface features like colour, organization and texture. Mazloom et al.~\cite{MazloomMultimodal2016} performed a multimodal study~(both text and image) on Instagram user posts and developed a popularity prediction model for brand-related posts. They found the visual and textual features to be complementary in predicting the post popularity. Majumder et al.~\cite{Majumder2017} present a method to extract \textit{human personality} traits from stream-of-consciousness essays using a convolutional neural network (CNN). They also use different features like LIWC~\cite{liwc2010}, Mairesse~\cite{mairesse2007}, character-level features, and responsive patterns~\cite{Sun2018}.

\subsection{Improving brand consistency as a sentence ranking task}\label{sec:sent-rank-recommend-prior}
The `\textit{alert detection}' task of RepLab 2013~\cite{amigo2013overview} was to predict the priority of an entity-related topic, from among the three classes - alert, mildly relevant and unimportant, in decreasing order of priority. They use ``\textit{polarity, centrality and user's authority}" to determine these priority assignments. 
Malmi et al.~\cite{malmi2019encode} develop a sequence tagging approach that frames the text generation task as a text editing task and outperforms existing \textit{seq2seq} models in sentence-level tasks of sentence fusion and sentence splitting. Pavlick and Tetreault~\cite{pavlick2016empirical} developed a statistical model for predicting formality at a sentence level, in the context of online debate forums and undertook a large-scale annotation task~(6574 sentences with 301 annotators). Munigala et al.~\cite{munigala2018persuaide} develop an unsupervised, text-generation~(at sentence-level) pipeline for generating persuasive descriptions~(usually one sentence long) from a given set of keywords, in a fashion domain. 

\noindent \textbf{Summary.} 
In this work, we characterize the notion of brand personality based on Aaker's formalization. We improve over the existing brand personality detection baseline, which only uses LIWC as the feature set, by contributing four more task-specific linguistic features (see Section~\ref{sec:linguistic}). We perform a characterization study where we study the impact of different company attributes like rank, sector and brand activities like \textit{brand extensions}, on brand consistency. Maintaining brand consistency is one of the brand reinforcement strategies that help to maintain the strength and favorability of brand associations~\cite{Keller1999}. Finally, for developing the sentence-level recommender system to improve the brand consistency of the textual content of company webpage, we use certain aspects from the ``Alert Detection task of RepLab 2013''. Our proposed \textit{brand consistency} aspect plays an essential role in the recommendation task.

\section{Dataset, Annotation and Features}\label{sec:features}
In this section, we describe the process in which the dataset has been collected, cleaned and timestamped. We then elaborate some of the typical properties of the dataset, the process of annotation followed, and the features derived for the dataset. The annotation helps us create ground truth for the undertaken classification task while the feature extraction helps in building effective classifiers. 

\subsection{Dataset}\label{sec:dataset}
 We collect the Fortune 1000 companies' textual content from their official websites for the year of $2017$. We include the webpages that contain the following information --- {\em about the company}, {\em media releases}, and {\em blogs and communication} directed towards the customers. We then use the Scrapy\footnote{https://scrapy.org/} framework to perform an extensive crawl and filter the pages based on keyword-based inclusion and exclusion rules over the webpage URL. We only consider webpages that are directed towards the customers --- (i) mention their brand characteristics that include about, history, vision, commitment, who-we-are; (ii) informative content like blogs, media releases, investors, and newsroom. We use the following inclusion keywords that are ---  \textit{about, about-us, news, press, introduction, strength, investors, history, vision, benefits, commitment, people, why-choose-us, who-we-are, approach, media, blog, social}.  We remove the following type of webpages from the crawling activity --- (i) product pages~\footnote{The product pages contain descriptions of multiple products in a single page and is embedded in the webpage as dynamic content. This requires separate crawling scripts for each individual page for each company, which requires a significant human effort; it is difficult to scale to ~650 companies} like showroom, products, store; (ii) content not targeted explicitly for consumers like legal, policy, disclaimer. The exclusion keywords used are \textit{job, jcr\_content, events, legal, help, showroom, products, store, project, career, policy, disclaimer, report}. We start from the home page of each company and only consider papers within the same domain name. Websites that contain non-English text content is also removed. We parse the ASCII text content with the paragraph~($<p>... ASCII text...</p>$) HTML tags for a given web page and append them together, partitioned by a paragraph separator marker. The final dataset (MT\textsubscript{large}) comprises of textual content from 643 companies and 299481 corporate webpages.

\subsection{Static and dynamic pages}\label{sec:statdynformulation}
Each of the company webpages consists of static and dynamic contents. 
Static pages define the brand represented by a company as its mission, vision, and core values. The frequency of such posts is usually very less and is mostly posted during the website's launch. Dynamic pages are used to communicate with the public on an ongoing basis. These are regularly posted as blogs, reports, media or press releases and comments to investors. Therefore, $MT_{large}$ consist of static pages from 338 companies and dynamic pages from 643 companies.
The static pages create the impression of the product while the dynamic content should retain it to ensure consistency of the brand. For static and dynamic websites, we divide the previously mentioned inclusion keywords~(present in a webpage URL)

\begin{tcolorbox}
\noindent \textit{Static page keywords:} introduction~(34), about~(573), commitment~(45), people~(252), vision~(48), strength~(429), history~(1116), approach~(571), benefits~(930) \\

\noindent \textit{Dynamic page keywords:} media~(19203), blog~(36844), news~(92448), press~(52544), investors~(5837)
\end{tcolorbox}

\subsection{Extracting temporal information from web content}~\label{extractTempInfo}
We needed information about the timestamp when the company posted a dynamic web page to understand the content's temporal behaviour. $MT_{large}$ consists of 298112 dynamic posts covering 643 companies, from which we are able to obtain timeline information from about half of them, resulting in 140,337 web pages. On manual inspection, we notice that the timestamp's granularity is in days or weeks. 75.01\% of these posts contain information on the day-level, while the remaining posts have a one-year temporal granularity.  We only look at posts made between January 2000 and September 2017.

\subsection{Basic observations}
Here, we conduct certain data analysis to understand the nature of the dynamic webpages. 

\begin{figure*}[!ht]
\centering
\includegraphics[scale=0.5]{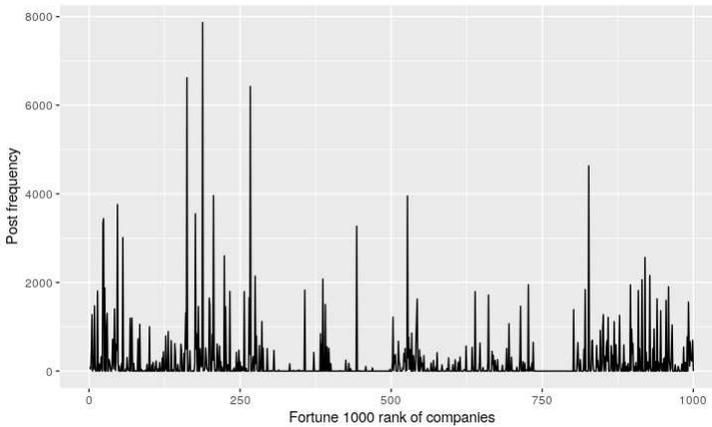}
\caption{Line plot of total posting volume per company.}
\label{fig:volumerankdyn}
\end{figure*}

\noindent \textit{Dynamic posting volume distribution.} Among the Fortune 1000 companies, we note that the number of posts is relatively identical (Figure~\ref{fig:volumerankdyn}), although we notice that there are occasional spikes reflecting firms that have a more-than-average posting volume. Higher ranked companies have more number of spikes.

\noindent \textit{Inter-arrival time between two dynamic posts. }
We examine the inter-arrival time between a single company's consecutive dynamic posts and then identify patterns that are common across multiple companies. By postings, we refer to dynamic web pages. Posts that have a granularity of only one day are considered, and this leads to 52432 posts.

\begin{figure*}[!ht]
    \centering
        \includegraphics[scale=0.5]{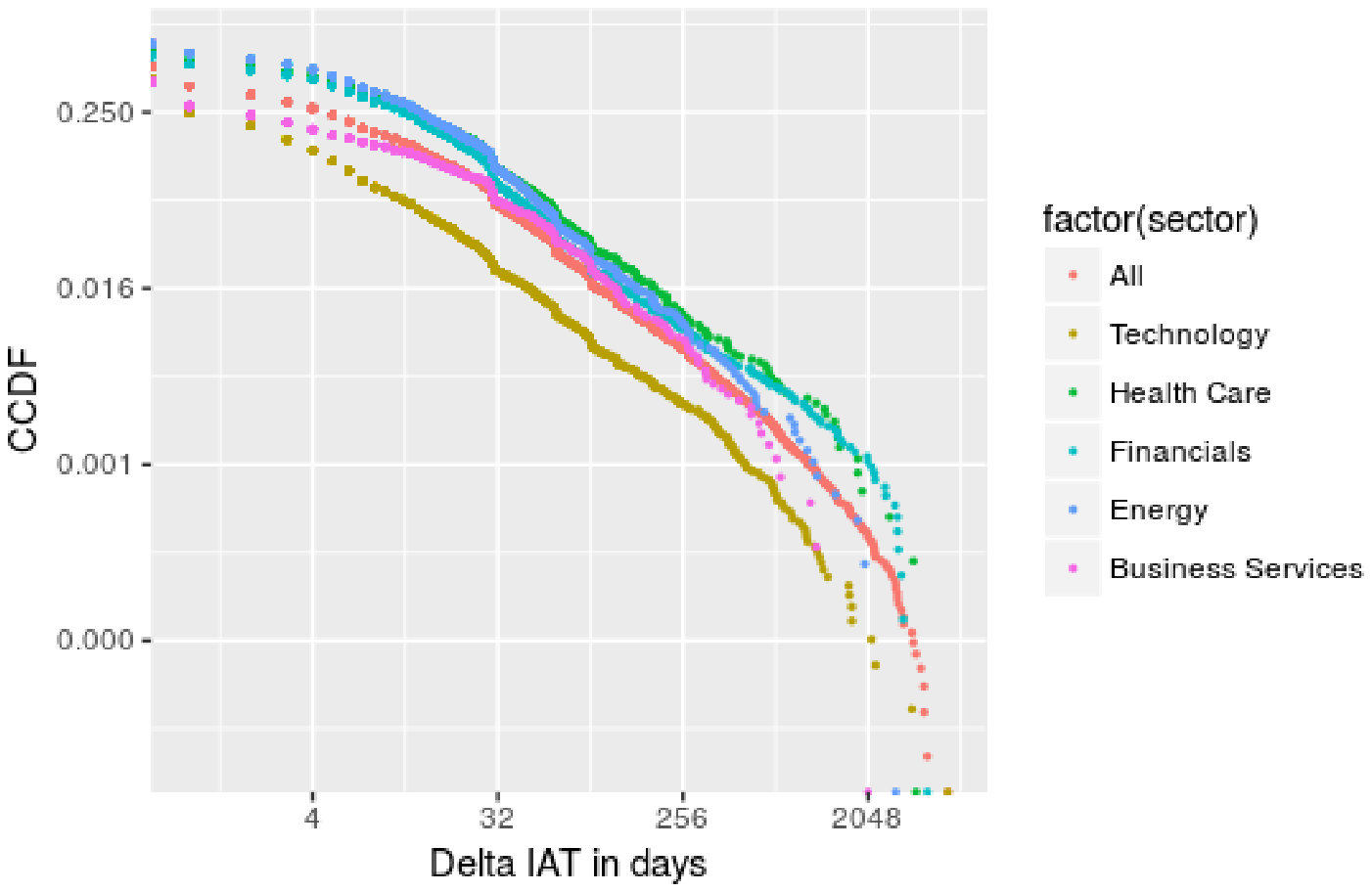}
    \hfill
        \includegraphics[scale=0.5]{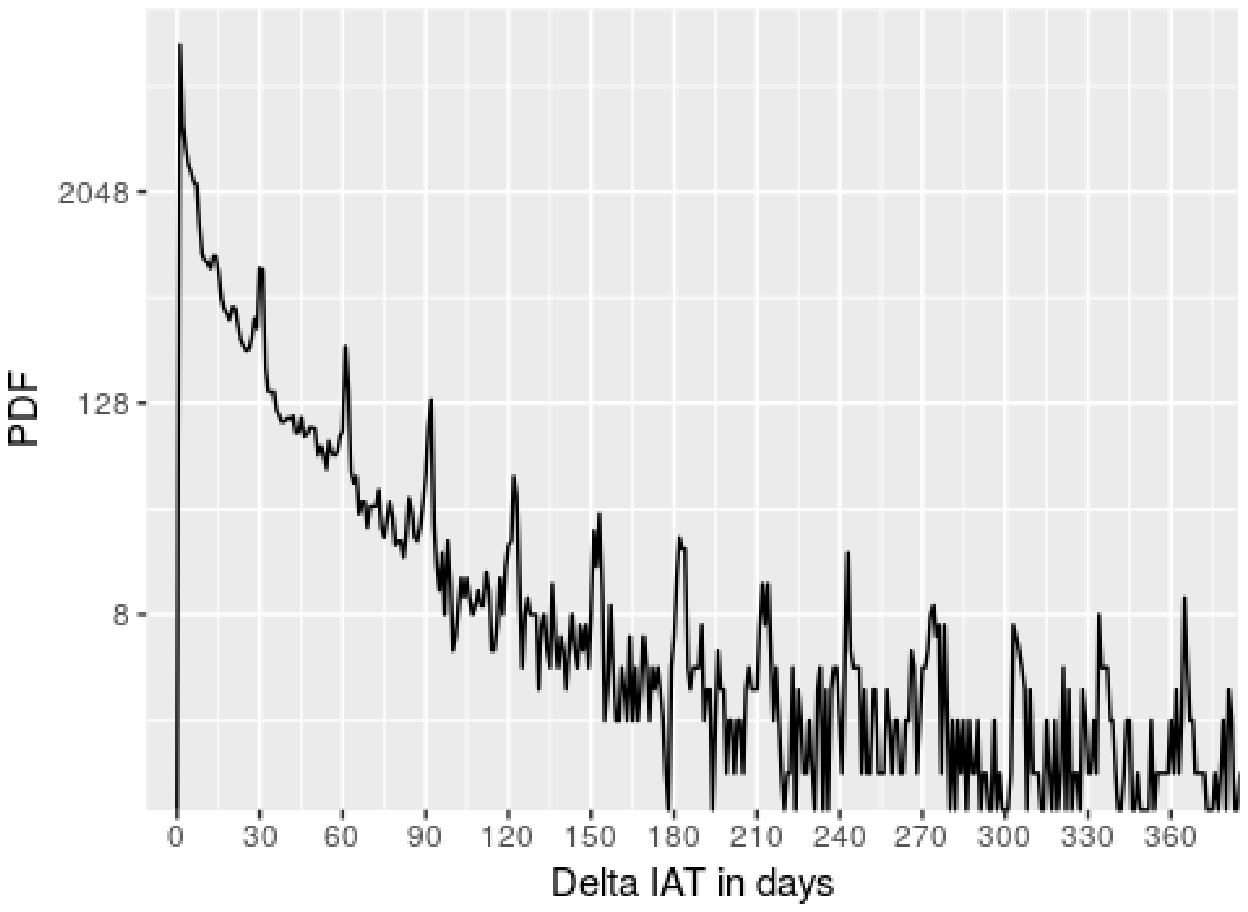}
    \caption{(left) Complementary Cumulative Distribution function (CCDF) plot of postings inter-arrival time~(IAT) over all companies. A heavy-tailed distribution is observed that is also maintained across the different sectors. (right) Periodic peak pattern: the Probability Density function (PDF) plot of posting's Delta inter-arrival time~(Delta IAT) in log base 2 scale, has peaked at 30 to 33 days interval. Delta IAT of a post is the posting time normalized by the posting time of the first web article  posted by the respective company.}
    \label{fig:iat_study}
\end{figure*}

Next, we conduct studies across different sectors and industries. We note that the inter-arrival time between two posts shows a heavy-tailed trend, which is also observed across other sectors~(in the left side of Figure~\ref{fig:iat_study}). We choose the top five sectors with the largest number of posts-- technology~(48219), financials~(11739), energy~(4915), healthcare~(4685) and business services~(3747), which also portrays a heavy-tailed behaviour. 
We consider the dynamic posts which have non-zero inter-arrival time and have day-level granularity. We detect peaks appearing consistently after an interval of 30 to 33 days~(in Figure~\ref{fig:iat_study} right side) and further investigate this issue, by studying what type of dynamic posts are more prevalent during these peaks. We note that at the month-end~\footnote{We designate `month-end' as the first two days and the last two days of a month.}, a majority (78\%) of them is posted. 
We observe that the dynamic post type that occurs most frequently is `news', among the other post types that are media~(9.63\%), blog~(13.98\%), news~(66.95\%), press~(25.67\%) and investor~(24.0\%).

\subsection{Annotation}\label{sec:tab:dataannotation}
Each article is annotated in two different ways --- (i) annotators adopt a five-point Likert scale to annotate the articles on each of the five dimensions of brand personality (similar to those used for annotating formality by~\cite{lahiri2015squinky} and \cite{pavlick2016empirical}). They are provided with an interface where the scores from 1 to 5 are present as a horizontal list of radio buttons from left to the right. There are five rows where each row corresponds to one of the five brand personality traits. 
(ii) they are then asked to rank the five dimensions in the same order that match the scores they previously provided (on a five-point Likert scale). Here, the annotator is asked to provide the rank in five boxes corresponding to each trait. 
In both cases, the order is shuffled to make the annotator more attentive.
We judge the fidelity of annotation based on the consistency between the two annotation methods. Each article is annotated three times by different annotators.

The complete interaction session of an annotator begins by getting familiarised with the brand personality traits. 
In order to familiarise an annotator with the brand personality traits, she is compulsorily made to spend some time to study the 
characteristics associated with each trait.
Examples of brands that best represent them are also provided for better comprehension. Then the annotator works for a session where she has to label three documents with a brand personality trait, followed by a test question of the form - \textit{What is domain or sector of the data point annotated just now? She may be engaged for multiple sessions.} 
 Prior knowledge of a brand or any personal experience may influence an annotator.
Thus to limit any external influences, while providing the document, we have replaced the organization name with a random string. Context to the annotator is provided by marking the domain or industry information. For example, we mark Merrill Lynch as a banking company. We use Amazon Mechanical Turk for crowdsourcing the text annotation of the randomly selected $600$ articles. We obtain articles that have scores from at least two annotators, after filtering using the consistency check as mentioned above. We only select the articles where at least two annotators have agreed, and this results in around 500 articles while the remaining 100 articles are discarded due to lack of guarantee or confidence by the annotator. When the absolute difference between the scores is equal or less than one, we consider that both the annotators have agreed on a given article. For example, resolving mismatch such as high sincerity may mean a score of 4 to one annotator and 5 to another annotator, which normalizes the biases between annotators. The inter-annotator agreement averages to 67.25\%, further outlined in Table \ref{tab:dataset}. We compute the mean of the annotator scores at this stage and use a static threshold of 3.0 to convert the score to a binary label. This indicates whether a particular brand personality is evoked from the text or not.

\begin{table}[!ht]
\centering
\begin{tabular}{ c|c|c|c  }

 Trait&Present&Absent&Agreement~(in \%)\\
 \hline
sincerity & 433 & 67&71.41\\
excitement & 339 & 161& 65.02\\
competence & 470 & 30& 75.62\\
ruggedness & 190 & 310& 63.50\\
sophistication & 276 & 224& 60.70\\

\end{tabular}
\caption{Inter-annotator agreement per dimension and the class distribution}
\label{tab:dataset}
\end{table} 

\subsection{Linguistic features}\label{sec:linguistic}
We use the concepts proposed by Delin et al.~\cite{delin2007} to formulate several linguistic features. We use them for capturing the trait of the underlying article in a more compact manner. We extract from each article the following set of linguistic features. \\
\begin{enumerate}[(i)]
\item \textbf{LIWC}: Linguistic Inquiry and Word Count \cite{liwc2010} is a dictionary of psycho-linguistics traits, that is widely used in psychological trait extraction and social data analysis~\cite{xu2012you}. For a given text, we use the values returned by the commercial API version~\footnote{www.receptiviti.ai} of LIWC.\\
    \item 
\textbf{Term Frequency-Inverse Document Frequency (TF-IDF)}: We first remove the stop words and then compute the TF-IDF vector for each document, which consists of unigrams, bigrams, and trigrams.\\
    \item 
\textbf{Contractions}: These are shortened phrase versions. Some examples include \verb+we are+ being substituted by \verb+we're+, \verb+is not+ being replaced by \verb+isn't+. A degree of informality and conversational tone is added to the text by contractions.\\
    \item \textbf{Collocations}: Collocations are word combinations that often occur together. Some examples are `heavy rain' and `high temperature'. We use the \verb+cite+ list of Pearson Academic Collocations, which consists of 2469 most common lexical collocations in written academic English, to construct our dictionary of collocations.\\
    \item \textbf{Chains of reference}: Chains of reference denote the brand's use of references to itself and closely associated elements. The key elements in this process are the noun phrases in the text, whose content and form can serve either to strongly evoke a brand, reinforce it, or not evoke it at all. Different kinds of relations that can hold between noun phrases and brand concepts as categorized by Delin et al.~\cite{delin2007} have been summarized in Table \ref{tab:chainref}. Here, we use repetition, partial repetition, co-reference, and possessive inferrable as four different features. \\
   \item \textbf{Readability}: This feature captures the simplicity of reading a given piece of text. The feature is based on the Flesch-Kincaid Readability Score~\cite{kincaid1975derivation}. The score considers the word length, sentence length, and the number of syllables per word. The higher the score, the easier it is to read. It is computed as follows:

$ Score=206.835-1.015\frac {TotalWords}{TotalSentences} - 84.6\frac {TotalSyllables}{TotalWords}$

\end{enumerate}

\begin{table}
\centering
\begin{tabular}{ p{2.5cm}|p{6cm}|p{3cm}  }
 
 Link & Definition& Example\\
 \hline
 Repetition & The full reference to the target company is referred to multiple times& Microsoft...Microsoft , Target...Target\\
 Partial Repetition & A reference to the target company is referred to by a phrase but refers to a concept that is not aligned with the target company & Microsoft...the Microsoft Service Promise, Target...Target Stores\\
 Co-reference & Full descriptive noun phrase is not used when a particular concept is again referred to & Microsoft...We, Target... With us\\
 Possessive inferrable & A possessive noun phrase is used when we create a link by referring to something that the company has, does, or has given to the customer & Microsoft...our network your phone, Target... our stores your cart\\
 
\end{tabular}
\caption{Various types of \textit{Chains of reference} relations that exist between a noun phrase and a brand concept, as categorized by Delin et al.~\cite{delin2007}.}
\label{tab:chainref}
\end{table}

\section{Classification Model}\label{sec:classifier-model}
\begin{figure}[!ht]
\centering
\includegraphics[scale=0.25]{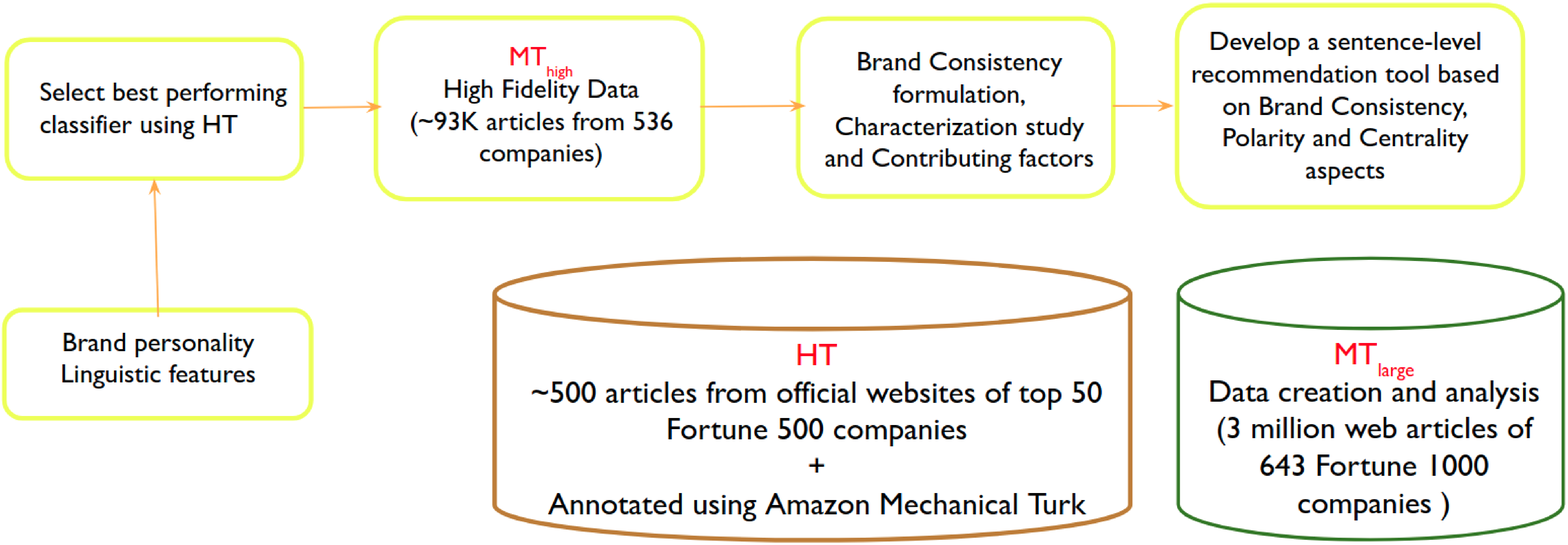}
\caption{Flowchart showing the methodological overview of our paper.}
\label{fig:method-overview}
\end{figure}

We train separate classifiers for the five traits independently. We provide a methodological overview of the paper in Figure~\ref{fig:method-overview}. 

We consider several classification models which include Naive Bayes, Logistic Regression and Decision Tree, Linear Support Vector Machines along with ensemble algorithms like Random Forest and AdaBoost. Although, most deep learning models require a large amount of labelled data to produce good results, certain models like fasttext are able to tackle low data volume and scarcity; we use it as a baseline model.
The human-annotated data is only fed to the classifier. The annotated data act as ground truth, while the features discussed in the previous section~(Section~\ref{sec:linguistic}) are fed into a classifier. During classification, we observe a significant class imbalance in the human-annotated data, as explained in Table \ref{tab:dataset}. We address the class imbalance problem by using a data-level approach called SMOTE~\cite{Chawla2002} where we increase the number of minority class data points (oversampling) to equalize the number of majority class data points in the training set. We use precision, recall and F1 score to evaluate the classification models. 
We obtain a 7-fold cross-validation strategy and compute our final score as the average of the obtained scores. 
Out of the classifiers trained, Linear SVM achieves the highest F1-score for all the brand personality traits, apart from competence (outlined in Table~\ref{tab:perfcompbaseline}). 
Fasttext, the only deep learning model, has high precision but very low recall value, and  is the second worst-performing model.
Since, linear SVM has the \textbf{highest F1-score of 0.802}, on average, after considering all the five traits, we select linear SVM as our final classifier.  As our feature set, we use LIWC~(which has 64 categories)~\cite{Xu2016}.

\begin{table*}[!ht]
\centering
\scalebox{0.72}{
\begin{tabular}{p{2cm} | c | c | c ||c|c|c||c|c|c||c|c|c||c|c|c}
Trait & \multicolumn{3}{c||}{sincerity}&\multicolumn{3}{c||}{excitement}&\multicolumn{3}{c||}{competence}&\multicolumn{3}{c||}{ruggedness}&\multicolumn{3}{c}{sophistication} \\ \cline{2-16}
 & Prec.&Rec.&F1& Prec.&Rec.&F1& Prec.&Rec.&F1& Prec.&Rec.&F1& Prec.&Rec.&F1\\ \hline
Naive Bayes &0.238&0.851&0.371&0.162&0.791&0.268&0.319&0.941&0.721&0.268&0.622 &0.319&0.141&0.854&0.239 \\
Logistic Regression &0.538&0.856&0.659&0.796&0.801&0.798&0.774&0.953&0.853&0.789&0.561 &0.654&0.733&0.727&0.725 \\
DecisionTree &0.774&0.871&0.819&0.661&0.742&0.698&0.921&0.954&0.937&0.568&0.536 &0.549&0.652&0.671&0.66 \\
RandomForest &0.838&0.865&0.85&0.72&0.795&0.754&0.953&0.939&\bf{0.946}&0.532&0.641&0.575 &0.623&0.737&0.673 \\
AdaBoost &0.85&0.868&0.859&0.746&0.761&0.753&0.923&0.95&0.936&0.606&0.585&0.589 &0.66&0.69&0.672 \\
SVM (Linear) &0.912&0.861&\bf{0.885}&0.832&0.801&\bf{0.815} &0.919&0.943&0.931&0.773&0.57&\bf{0.655}&0.751&0.707&\bf{0.725}\\
fasttext&0.86&0.32&0.46&0.83&0.27&0.41&0.9&0.15&0.26&0.61&0.33&0.43&0.81&0.4&0.53\\
\end{tabular}}
\caption{Performance comparison for selecting the optimal classifier among the various binary classification algorithms. Linear SVM performs the best for all the brand personality dimensions, except for competence.}
\label{tab:perfcompbaseline}
\end{table*}

Next, we add different linguistic features to the empirically determined feature set of LIWC. We incrementally expand the feature sets one by one and finally select the optimal set of features which give the highest F1-score. We obtain a very high F1-score for the brand personality detection task for sincerity, excitement and competence. However, for the remaining two traits (ruggedness and sophistication), we achieve a high precision value of 0.773 and 0.751, respectively. We outperform the (limited-feature-set) linear SVM by 2.49\%, achieving an F1-score of 0.822 when the best feature set for each trait is utilized. We are able to achieve \textbf{an F1-score of 0.822}, which is an improvement over the (limited-feature-set) linear SVM by 2.49\%.

We further validate our model by utilizing the ground-truth in which the annotators have provided the ranks of the five brand personality traits in order of their presence in the provided text as well as individual trait-wise score. We use the respective classifiers to obtain the trait-wise confidence score for each text and then create an ordering among the five traits. Pearson correlation coefficient is then computed between the individual trait-wise scores provided by the annotators with the confidence score provided by the classifiers. Spearman rank correlation coefficient is then computed between the ranks provided by the annotators and the rank that we obtain based on the confidence scores. The improvement (deterioration) after the feature addition step shows high correlation with both the Pearson and the Spearman score (explained in Table~\ref{tab:oc_comp}). More importantly, we observe that the classifier set formed using the best feature set (which is different for different traits), also achieves the best performance in both of these metrics~(Pearson - 0.426, Spearman - 0.431).

We call the set of five classifiers with the best performing features (unique for each trait) model as FLCS (Final Linear Classifier Set). FLCS is used to annotate MT\textsubscript{large}, for performing the brand consistency study~(covered in next section). 

\begin{table*}[!ht]
\centering
\scalebox{0.65}{
\begin{tabular}{p{1.5cm} | c | c | c ||c|c|c||c|c|c||c|c|c||c|c|c||c|c}

Trait & \multicolumn{3}{c||}{sincerity}&\multicolumn{3}{c||}{excitement}&\multicolumn{3}{c||}{competence}&\multicolumn{3}{c||}{ruggedness}&\multicolumn{3}{c||}{sophistication}& PMax & SpMax \\ \cline{2-16}
 & Prec.&Rec.&F1& Prec.&Rec.&F1& Prec.&Rec.&F1& Prec.&Rec.&F1& Prec.&Rec.&F1& & \\ \hline
liwc (baseline) &0.912&0.861&0.885&0.832&0.801&0.815&0.919&0.943&0.931&0.773&0.57&\bf{0.655}&0.751&0.707&\bf{0.725}&0.409&0.372 \\
liwc+ tfidf  &0.988&0.867&0.923&0.9&0.787&\bf{0.839}&0.998&0.94&\bf{0.968}&0.527&0.573&0.545&0.726&0.704&0.707&0.407&0.429 \\
tfidf*+ contractions &0.989&0.867&0.923&0.894&0.784&0.834&0.998&0.94&0.968 &0.527&0.578&0.548&0.725&0.7&0.708&0.406&0.424 \\
cont*+ collocations &0.989&0.867&0.923&0.897&0.787&0.837&0.998&0.94 &0.968&0.522&0.579&0.545&0.722&0.708&0.709&0.405 &0.428\\
coll*+ chainref &0.991&0.867&\bf{0.925}&0.894&0.787&0.836&0.998&0.94&0.968&0.559&0.587 &0.569&0.729&0.693&0.706&0.404 &0.407 \\
chainref*+ readability&0.989&0.868&0.924&0.862&0.805&0.837&0.998 &0.94&0.968&0.59&0.6&0.592&0.722&0.718&0.72&0.406&0.419 \\ 
Best features (FLCS)&0.991&0.867&0.925&0.9&0.787&0.839 &0.998&0.94&0.968&0.773&0.57&0.655&0.751&0.707&0.725&\bf{0.426}&\bf{0.431}\\
\end{tabular}}
\caption{Performance comparison across different feature sets and formation of the Final Linear Classifier Set (FLCS). FLCS improves over the (limited-feature-set) linear SVM by 2.49\%, and achieves a F1-score of 0.822. FLCS correlates strongly with the rank annotations (Pearson and Spearman rank correlation coefficient).}
\label{tab:oc_comp}
\end{table*}

\subsection{High fidelity points}\label{sec:HFPT}
A requirement for carrying out brand consistency is the availability of a large number of labelled data points. Therefore, we use FLCS to classify the MT\textsubscript{large} data and conduct the brand consistency study with the data points that are classified with high confidence ($\geq$ 0.095) in at least one of the five classifiers. We check whether the selected set of points actually agree with the class for which it is annotated, through manual cross-checking. We randomly checked 50 data points and observed only two errors. This dataset is named as MT\textsubscript{high}. The different data collection steps are described in Table~\ref{tab:datacollectionsummary}.

\begin{table}[!ht]
\centering
\begin{tabular}{p{2cm}|p{1.5cm}|p{1.5cm}|p{6cm}}
\hline
     Dataset name& Number of posts&Number of companies&Collection strategy\\\hline
    $MT_{large}$ & 298112 & 643 & Web scraping from official websites based on accept and deny keywords \\ \hline
    HT & 500 & - & Randomly selected 600 points from $MT_{large}$, which satisfy strict annotation criteria\\ \hline
    $MT_{high}$ & 93321 & 536 & Subset of $MT_{large}$, which is annotated with high confidence by FLCS\\ \hline
    $MT_{time}$ & 49833 & 242 & Subset of $MT_{high}$ having timestamp data\\ \hline
    $MT_{noTime}$ & 43488 & 512 & Subset of $MT_{high}$ without having timestamp data\\ \hline
    Inter-brand relations & - & 386 & Companies mentioned in at least five web articles\\ \hline
    Product promotions & 3255 & - & Subset of $MT_{high}$ identified as product promotion posts\\ \hline
    Brand extensions & 977 & - & Subset of $MT_{high}$ having at least one brand extension mention\\ \hline
    $MT_{auth}$ & 89169 & - & Subset of $MT_{high}$ after removing product promotions and brand extension posts\\ \hline
    $MT_{study}$ & 202 & - & Subset of $MT_{high}$ used for sentence recommendation task\\ \hline
    $HT_{recco}$ & 36 & - & Subset of $MT_{study}$ that is manually annotated to form the ground-truth for the recommendation task\\ 
\end{tabular}
\caption{Summary of the different data collection steps.}
\label{tab:datacollectionsummary}
\end{table}

\section{Brand consistency characterization study}\label{brandconsanalysis}

We investigate how well a company exhibits and maintains its brand personality within the web content produced over multiple content categories and across time. 
As per our knowledge, this is the first study which investigates the notion of brand consistency computationally. To conduct the study, we first define consistency between two web page content and extend it to a company-level metric. Next, we study the effect of topical consistency of static posts, product promotion posts, inter-brand relations and brand extensions on brand consistency. Finally, we formulate a temporal brand consistency score and study the role played by the company's ranking.

\subsection{Brand consistency - Formulation}\label{sec:brandconsformulation}
We represent each post by two five-dimensional vectors --- {\em label vector} and {\em rank vector}. 
`Label vector' is used to indicate whether a trait is either present or absent from the text and is represented as a binary label. `Rank vector' stores the position~(rank) for each brand personality trait, when they are arranged in decreasing order of their degree of presence in a text. We compute the degree of presence based on its confidence score and accordingly determine the order. 
The similarity between a post and the representative vectors (static post) of the respective company is calculated using two measures -  $binLabelSim$ and $rankLabelSim$. The most frequently occurring label and rank vector among all the static posts is considered as the respective representative vectors~(stands for the company's brand personality).

\begin{enumerate}[(i)]
    \item \textit{binLabelSim} : The Hamming and Levenshtein distances between two label vectors  are used to compute a composite score termed as \textit{binLabelDist}. It is computed as the average value of their Hamming and Levenshtein distance measures, where \textit{binLabelSim} is 1 - \textit{binLabelDist}.
    \item \textit{rankLabelSim}:   Pearson, Kendall Tau and Spearman's rank correlation coefficients are used to compute the similarity between two such rank vectors. rankLabelSim is the mean value of Pearson, Spearman and Kendall tau's rank correlation scores. Spearman and Kendall's tau rank correlation coefficients compute the correlation between the respective rank vectors while for computing the Pearson correlation, we directly use the scaled confidence score from each of the trait-specific classifiers.
\end{enumerate}

We first check the level of consistency of static posts with the formulated metrics.
The value of `binLabelSim' and `rankLabelSim' for static posts is $0.94 \pm 0.13$ and $0.86 \pm 0.29$ respectively, which indicates a very high degree consistency; the result in turn also demonstrate the efficacy of the metric.  For dynamic posts, `binLabelSim' is $0.65$ and `rankLabelSim' is $-0.03$ respectively, with a high standard deviation, which indicates that the brand consistency among such posts is largely missing.  It is evident that \textit{rankLabelSim} is more sensitive than \textit{binLabelSim}, and the value can vary heavily with even a small change in the scores of individual traits. Therefore, rankLabelSim is  used only 
as a measure to break a tie in binLabelSim.

\noindent{\bf Consistency levels :} The different brand consistency levels are named as `strongly consistent, consistent, and not consistent' (see Table \ref{tab:constlevels}). The conditions are determined through manual inspection. We have learnt from inspection that \textit{binLabelSim}, play a more determining role in reflecting the different categories of the brand consistency levels formulation. 
\textit{rankLabelSim} only act as a secondary measure. 

\begin{table}[!ht]
\centering
\begin{tabular}{ c|c|c}
\hline
 Brand consistency level&binLabelSim&rankLabelSim  \\
 \hline
strongly consistent &$\geq 0.8$ & $\geq 0.6$\\
consistent & Otherwise & Otherwise \\
not consistent & $\leq 0.5$ & $\leq 0.6$ \\ \hline
\end{tabular}
\caption{The three brand consistency levels and its associated conditions in terms of binLabelSim and rankLabelSim; the strict ordering for rankLabelSim is not maintained as it is used as a secondary measure. }
\label{tab:constlevels}
\end{table} 

\noindent {\bf Consistency score~(BrandConsScr) :} It is defined as the fraction of the total number of posts 
that are consistent or strongly consistent with the static post set.

\subsection{Brand Consistency of Organizations}
\begin{figure}[!ht]
\centering
\includegraphics[scale=0.5]{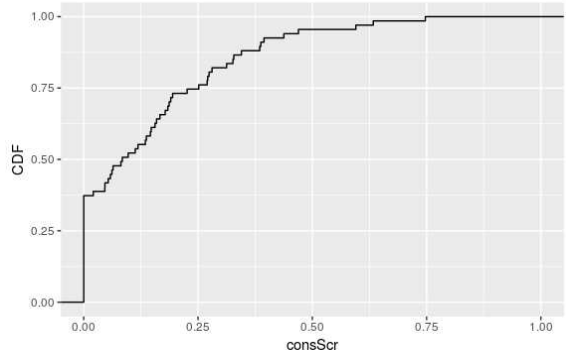}
\caption{Cumulative Distribution Function~(CDF) over the consistency scores for all the companies. Most of the companies are not able to maintain a decent brand consistency score.}
\label{fig:cdfConsScr}
\end{figure}

We demonstrate the relationship between the consistency score and the fraction of companies that are able to maintain that brand consistency score in Figure~\ref{fig:cdfConsScr}. We only consider organizations that contain at least one static post, with at least one trait being positive (present). We observe that a high consistency score is achieved only by a few companies, which demonstrate the significance of the current work.  

\subsubsection{\bf Companies with high  brand consistency}\label{sec:timeinvariantfactors}
Here, we measure the ability of a company to maintain high brand consistency across the posts made by the company.
We compute the percentage of \textit{strongly consistent} company postings among the total posting volume. We mention the five companies that have the highest percentage of strongly consistent posts in Table~\ref{tab:constdyn1}. A very high mean value (binLabelSim: 0.8, rankLableSim: 0.7) and very low standard deviation (binLabelSim: 0.08, rankLableSim: 0.07) is observed for the top five companies. We measure the percentage of posts instead of their absolute count and thus, only consider companies having at least twenty strongly consistent posts.

\begin{table}[!ht]
\centering
\scalebox{0.9}{
\begin{tabular}{ p{2.75cm}|p{3cm}|p{1.75cm}|p{1.75cm}|p{2.0cm}|p{2.0cm} }
\hline
 Company name &Ratio of strongly consistent posts~(in \%) &binLabelSim (mean) &binLabelSim (sd) &rankLabelSim (mean)&ranklabelSim (sd)\\
 \hline
 FTI Consulting Inc. & 1.0&0.9&0.1 & 0.68 & 0.08\\
 Regis Corporation & 0.54&0.88&0.1 & 0.67 & 0.06\\
Engility Holdings Inc.&0.42&0.84 & 0.08 &0.66 & 0.05\\
Caesars Entertainment Corporation& 0.41& 0.8&0.0  & 0.81 & 0.1\\
Prudential Financial Inc. & 0.23&0.91&0.1 & 0.68 & 0.08\\
\hline
\end{tabular}}
\caption{Top five companies with highest percentage of strongly consistent posts.}
\label{tab:constdyn1}
\end{table} 

\subsubsection{\bf Sector-wide analysis} 
We expand the search and investigate the notion of brand consistency sector-wise.  Here we study the following sectors - technology, healthcare, financial, energy, business services and media. We compute the BrandConsScr score of a company using $MT_{high}$ data points and show their performance across the sectors in Table~\ref{tab:sectorConsLevels}. We observe that the number of companies present in each sector is unevenly distributed, and thus ranking the sectors by simply averaging over all the company consistency scores, may not be a good measure. We thus rank each sector based on the count of companies, when binned in terms of their `BrandConsScr' score (four bins are made --- bin 1: [$\geq 0.7$], bin 2: [0.5 to 0.7], bin 3: [0.3 to 0.5], bin 4: [$\leq 0.3$]). Instead of computing the brand consistency score, we only return a ranked list, stable-sorted in non-increasing order of bin 1, followed by bin 2 to bin 4. We observe that the `financial' sector has the best performance while `media' is showing the least consistency. This is mainly because the `media' companies have to cover a more diverse and complex range of topics (both domestic and international), which makes it more challenging to maintain a consistent brand personality in their content.

\begin{table}[!ht]
\centering
\scalebox{0.75}{
\begin{tabular}{ p{2cm}|p{1.5cm}|p{1cm}|p{1.5cm}|p{1.5cm}|p{2cm}|p{2cm}|p{2cm}|p{2cm}}
 \hline
 Sector& Total companies & Total posts& ConsScr mean&ConsScr maximum & companies count (ConsScr $\geq 0.7$) & companies count (ConsScr $\geq 0.5$)&companies count (ConsScr $\geq 0.3$)&companies count (ConsScr $\leq 0.3$)\\
 \hline
 Financials & 26& 6221& 0.35&0.98& 4& 4&3&15\\
Energy & 19& 2434&  0.25&0.74& 2& 1&3&13\\
Technology & 21& 18833& 0.23&0.83& 2& 0&3&16\\
Health Care & 9& 1223& 0.32&0.79& 2& 0&3&4\\
Business Services & 3& 674 & 0.42&0.85& 1& 0&1&1\\
Media & 2& 626 & 0.24&0.29& 0& 0&0&2\\
 \hline
\end{tabular}}
\caption{Brand Consistency performance across the different sectors. The `Financials' sector performs the best while `Media' is performing poorly. We shorten \textit{BrandConsScr} to \textit{ConsScr} due to space-constraints.}
\label{tab:sectorConsLevels}
\end{table} 

\subsection{Nature of static and dynamic post}\label{sec:static-dyn-posts}
As previously defined in Section~\ref{sec:statdynformulation}, static posts represent the mission and vision of a company. 
According to our brand consistency formulation, the static posts form a representative vector, signifying the target brand personality of the respective organization that each dynamic post wants to achieve. Here we study the different factors that may affect the quality of brand consistency prediction.
\subsubsection{\bf Topical consistency among the static posts}
We investigate whether companies that have low topical consistency among their static posts face more difficulty in maintaining brand consistency among its webpage postings. We randomly select $50000$ documents from $MT_{high}$ and further train a Latent Dirichlet Allocation~(LDA) topic model on this set of documents. We choose the number of topics to be $100$ and number of passes as $10$. 

\begin{tcolorbox}
\noindent \textit{Topics having multiple companies from the same sector}: health, care, service, patient, medical, healthcare, blue, \textit{davita, cigna, wellpoint}; \textit{boeing}, see, lake, angeles, los, \textit{denver}, conference, seattle, activist, airplane

\noindent \textit{Single-brand topics}: \textit{cypress}, product, semiconductor, memory, manufacturing, flash, design, device, pc; airline, aircraft, flight, service, company, customer, \textit{cessna}, world, aviation
\end{tcolorbox}

Given that we have a trained LDA topic model from the previous step, we obtain the topic-wise scores for all static pages. For each company, we compute a topical consistency score~(TopicConsScr) as the mean pair-wise cosine similarity score among each static post pair.
We consider companies having at least $2$ static posts, which leads to $79$ companies. We observe that the BrandConsScr score for companies with low topical consistency~(TopicConsScr $\leq 0.5$) is significantly lower than companies with high topical consistency~(TopicConsScr $> 0.5$) in terms of mean~($0.157$ vs. $0.335$) and maximum~($0.65$ vs. $0.983$) value of BrandConsScr score of a company. This may indicate that it is more difficult to maintain brand consistency for companies with low topical consistency among their static posts. Table~\ref{tab:topic-cons-comp} outlines the list of companies in terms of the extreme values of TopicConsScr and BrandConsScr.

\begin{table}[!ht]
\centering
\begin{tabular}{ p{2cm}| p{2cm} |p{8cm}}
\hline
 TopicConsScr & BrandConsScr & Company name\\
 \hline
 High & High & Ally Financial Inc., One main Holdings, Inc.\\ \hline
 Low & Low & Intel Corporation, Akamai Technologies, Inc., Zebra Technologies Corporation, Genesee \& Wyoming Inc., SL Green Realty Corp., Herman Miller, Inc., Sears Holdings Corporation\\ \hline
 Low & High & None\\ \hline
 High & Low & Securian Financial Group, Inc., The Blackstone Group L. P., Spectra Energy Corp, Freeport-McMoRan Inc., American Greetings Corporation\\ \hline
\end{tabular}
\caption{List of companies in terms of extreme values of topical consistency~(TopicConsScr) and brand consistency score (BrandConsScr)}
\label{tab:topic-cons-comp}
\end{table} 


\subsubsection{\bf Product promotion posts}\label{sec:brandinvariantfactors}
Organizations often post promotional content to boost event outreach and product popularity among consumers.
We observe competence to be the primary trait (72\%) for such promotion-related posts.
 A trait is primary for a given post if it is ranked first or second among the five traits.

In order to study its effect on the brand consistency score of a company, we remove these promotion-specific posts (see Table~\ref{tab:datacollectionsummary}) from $MT_{high}$ and recompute the consistency scores as explained in Section~\ref{sec:timeinvariantfactors}.
We observe a high positive correlation of $0.803$ between the proportion of promotion posts and improvement in BrandConsScr score after removing the promotion posts. 
We also observe a medium degree of positive Pearson correlation of $0.45$ (statistically significant) between rank improvement due to the removal of the product promotion posts and the percentage of company posts which are promotions. That is, on removing the promotion posts not only the absolute consistency increases but also companies which can shed a higher fraction of promotion post can go higher up in the rank ladder.
The study indicates that although a company may be deliberately trying to portray a competent image during promotional events which may have an adverse impact in their overall brand image. 

\subsection{Inter-brand relations and Brand Extension}\label{sec:interbrand}
We have till now assumed an organization to represent a homogeneous set of products~(portraying the same brand personality). However, in a real-life scenario,  especially large organizations, interact and develop relationships with other brands and organizations. These interactions may be explicit in nature, like launching diverse product lines (brand extensions) as well as implicit like forming partnerships and collaborations with other organizations (inter-brand relations). We study the impact of such high-level interactions on brand consistency.

\subsubsection{\bf Brand extension}\label{sec:brandextension}

A single company may have multiple brands~(known as umbrella branding) or may have multiple product lines~(known as the brand extension). There exist three dimensions to measure the fit of extension - the `complement'~(satisfying the consumer's specific needs), the `substitute'~(products having the same user situation for satisfying the same needs), and the `transfer'~(describes the brand's perceived ability and their extension products)~\cite{aaker1990brand}. 
We investigated whether brand dilution takes place due to brand extension, for that we manually check and finally select 10 {\em Fortune 1000 companies} that exhibit brand extensions. We observe that the average consistency score of these companies is around $5\%$ lower than the mean computed across all the companies.
We, therefore, manually identify mentions that we consider as brand extension variants - (i) assets and subsidiary company~(CVS Health); (ii) various company branches in different countries~(Fujitsu); (iii) research division~(Gartner Research) and removed all the posts having at least one such mention~(see Table~\ref{tab:datacollectionsummary}). The median consistency score of these ten companies improved by $4.31\%$~(mean improved by $0.6\%$), which may indicate that presence of brand extension-related posts negatively (though marginally) affects the brand consistency of a company to certain extent.

\begin{tcolorbox}
\noindent \textit{Complementary products}: Colgate paste and Colgate brush; Kodak batteries and Kodak cameras; Dura cells and Dura beam flashlights

\noindent \textit{Substitute products}: Oreo cakes, Oreo cookies, and Oreo ice creams; Nescafe coffee, Nescafe chocolates, and Nescafe cold coffee
\end{tcolorbox}

\subsubsection{\bf Inter Brand Relation}
We study the interaction of a brand with other brands, which helps the consumers to identify and connect with a brand at a more social level~\cite{schmitt2012consumer}.
 We extract phrases tagged as `organization' using Stanford Named Entity Recognizer~(NER) Tagger~\cite{finkel2005incorporating}, from all the documents of $MT_{high}$. We consider \textit{organization mentions} that have a partnership or collaboration with the respective Fortune 1000 company and occur in at least five of its web articles --- (i) \textit{Akamai Technologies, Inc.} : Turner Sports, Microsoft Silverlight, Cisco, Netflix, Apple, Adobe, eBay; (ii) \textit{The Goodyear Tire \& Rubber Company} : ESPN, NASCAR, Ford Motor Company, Credit Agricole Securities (USA) Inc., PricewaterhouseCoopers LLP
 
 We observe that the presence of inter-brand relations may negatively impact brand consistency to certain degree only when it exceeds a threshold number of organizations (in our case, it is around 50). The mean consistency score of these companies drops to 0.21, which is 20.1\% lower than the average brand consistency score. This is because companies usually need to interact with other organizations to perform their day-to-day activities. However, it becomes difficult to maintain the same level of consistency when the number of organizations rises significantly.

\subsection{Temporal brand consistency characterization study}

The study of brand consistency till now has not considered the temporal aspect that is how a company performs as time passes from their initial website launch. 
We use our previously formulated brand consistency levels and group posts for every 12 weeks duration. The time period of 12 weeks is arbitrarily determined based on the sparsity of the available data points. Five companies with the highest number of consistent temporal bins is outlined in Table~\ref{tab:cntDisConstBins}. We consider a given temporal bin for a company to be consistent if BrandConsScr $\geq 0.5$.
We observe that the companies are all over the places in terms of rank, and none of the top-ranked companies is not able to enter the list. This may be due to the presence of a wide variety of product lines, branches in different countries, and elaborate marketing campaigns.  However, even though the top-ranking companies may not reach the podium, it may so happen that on an average, they maintain better consistency. This is what we investigate next. 

\begin{table}[!ht]
\centering
\begin{tabular}{ p{4.5cm}| p{1.5cm}| p{1.5cm}|p{1.5cm}|p{2.5cm} }
\hline
Company name& Fortune 1000 rank & ConsScr mean& ConsScr sd&Total no. of consistent bins\\ \hline
Capital One Financial Corporation & 100 & 0.44& 0.24 & 45\\
Engility Holdings, Inc. & 915& 0.75& 0.15 & 11\\
Westlake Chemical Corporation & 507  & 0.47 & 0.32 & 11\\
Regis Corporation & 1000 & 0.63&0.17 & 4\\
Principal Financial Group, Inc. & 227 & 0.6 & 0.15 & 4\\
\hline
\end{tabular}
\caption{Top five companies ranked in terms of total number of consistent temporal bins and the mean value of brand consistency score (BrandConsScr). We shorten BrandConsScr to ConsScr due to space-constraints.}
\label{tab:cntDisConstBins}
\end{table} 

\subsubsection{\bf Relation between ranking and brand consistency }
\begin{figure}[!ht]
\centering
\includegraphics[scale=0.6]{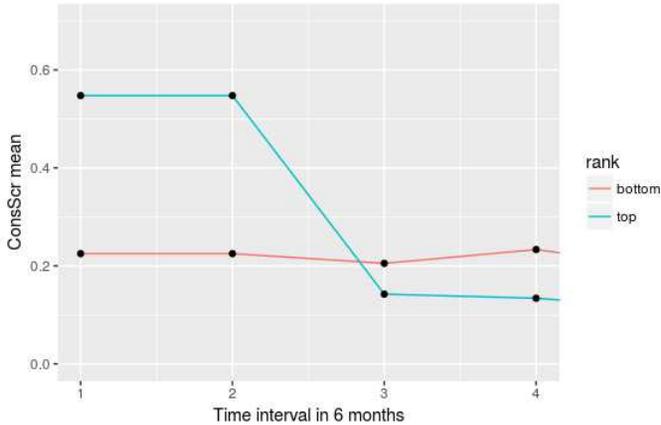}
\caption{Performance comparison between top and bottom ranked company for brand consistency over half year timestep.}
\label{fig:consScrMean}
\end{figure}

We study whether there exists any relationship between its rank in the Fortune 1000 list and a company's ability to maintain brand consistency.
We consider the top-ranked companies as those companies that are ranked within 150 of the Fortune 1000 companies, and the bottom-ranked companies as those companies that are ranked between 850 and 1000. We restrict our analysis to include only those companies that have at least 25 dynamic web pages. Finally, we have 18 top-ranked companies and 20 bottom-ranked companies. Instead of the previous temporal bin duration of 12 weeks, we perform temporal binning of 6 months duration. This is done because the number of data points per temporal bin, was significantly sparse. 
We observe that during the first year, the top-ranked companies are able to maintain a higher average consistency score when compared with the bottom-ranked companies. However, the performance of the top-ranked companies drops to a consistency score which is equal to the bottom-ranked companies~(see Figure~\ref{fig:consScrMean}). The slight difference between the two, which is seen from 18 months is not statistically significant. 
A plausible reason behind the higher consistency maintained by the richer companies can be attributed to the employment of expert and costly human resources by these companies to manage the content. 
There can be various reasons behind the drop beyond eighteen months, one probable reason 
being a  company slowly repositions  its brand as time passes. 
\begin{tcolorbox}
\noindent \textbf{In terms of dynamic posts count, the top five companies}: Microsoft Corporation~(5365), Bank of America Corporation~(467), Intel Corporation~(294), Capital One Financial Corporation~(282), Starbucks Corporation~(143)

\noindent \textbf{Bottom five companies}: Red Hat, Inc.~(732), Autodesk, Inc.~(270), Engility Holdings, Inc.~(225), Akamai Technologies, Inc.~(180), Overstock.com, Inc.~(128)
\end{tcolorbox}

\section{Sentence ranking tool} \label{recommendertool}
In this section, we develop a {sentence ranking} system. {More formally, given a document that is not consistent with the desired brand personality, we develop a helper tool that identifies the top three inconsistent sentences which can then be corrected or modified (by a human expert)  to improve the brand consistency of the document.}

\subsection{Proposed methodology}
The sentences are ranked based on three different aspects. Apart from brand consistency, we consider the aspects of negative polarity and centrality explored in the ``Reputation Alert Detection'' task of RepLab 2013~\cite{amigo2013overview}. 
We obtain $4668$ sentences from the articles selected for this ranking task~(see Section~\ref{sec:recco-data-annot}). 

\noindent \textit{(i) Negative polarity~(POL): }
We use a lexicon and a rule-based sentiment analysis of NLTK toolkit, called \textit{VADER Sentiment Analyzer }~\cite{HuttoVader2014} and label a sentence as `negative' (whetherNeg) if its negative score~(negScr) is higher than its positive sentiment score. Out of $4668$ sentences, only $8.2\%$ have a negative sentiment word in a sentence. 

\noindent \textit{(ii) Centrality~(CTR): }
 We measure the \textit{centrality} of a sentence in the following manner --- {We  first identify the 
 entities (generally, a noun phrase) present in a document 
using the popular Spacy\footnote{\url{spacy.io}} tool and performing certain reference resolution (of a pronoun to the corresponding noun). 
We then consider only those entities which have occurred at least three times and manually shortlist  a list of `central entities' 
(entities that are related to the target organization). 
We mark a sentence with \textit{whetherCentral} score equals to $1$ if the sentence contains at least one of the `central entities', otherwise we mark it as $0$. }

\noindent \textit{(iii) Brand consistency~(CONS): } The brand consistency of a sentence is measured using method explained in Section~\ref{sec:brandconsformulation}.  Here we measure inconsistency based on 
$binLabelSim$ which assumes the values [0, 0.2, 0.4, 0.6, 0.8, 1].

\noindent {\bf Sentence scoring : } 
Given the aspect-wise scores of each sentence, 
we propose a relevance score, in the scale of 1 to 6; based on a combination of the three criteria calculated \textit{whetherNeg~(binary)}, \textit{whetherCentral~(binary)}, \textit{binLabelSim} values~(see Table~\ref{tab:all3formula} for details). 

\noindent {\bf Sentence ranking :} The {\em Multi-Aspect Sentence Ranking} model (called `MASR-3') uses the sentence-wise aspect-specific scores in the descending order of relevance score, if there is a tie then descending order of \textit{negScr}, and subsequently descending order of target entity size is considered.

\begin{table}[!ht]
\centering
\begin{tabular}{ c|c|c|c}
\hline
 Relevance score&whetherNeg&whetherCentral&binLabelSim \\ \hline
6 & yes & yes & - \\
5 & yes & no & - \\
4 & no & {yes} & $\leq 0.5$ \\
3 & no & {yes} & $> 0.5$ \\
2 & no & {no} & $\leq 0.5$ \\
1 & no & {no} & $> 0.5$ \\ \hline
\end{tabular}
\caption{Sentence scoring function of MASR-3 model, which outputs a relevance score, in the scale of 1 to 6. }
\label{tab:all3formula}
\end{table} 

\subsection{Dataset and ground-truth creation}\label{sec:recco-data-annot}
At first, we 
select the `not consistent', moderate length (300 to 500 words) web articles 
that are sufficiently readable (\textit{Flesch's Readability Ease} score~\cite{kincaid1975derivation}  greater than 20) from the MT\textsubscript{high} posts {(see Table~\ref{tab:datacollectionsummary})}. 
We further remove the 
articles that do not contain at least one emotionally-charged words, to avoid articles that are explicitly neutral in nature. 
We observe that Microsoft and Nordstrom have a significantly higher number of `not consistent' posts due to its overall high posting volume. We remove all the articles from both the companies to prevent our observations from being heavily influenced by them.  Thus finally we obtain a set of 202 posts which we term as    MT\textsubscript{study}.
We randomly select 36 articles from MT\textsubscript{study} and use two annotators\footnote{Both the annotators are well-versed in English, and are not authors of the paper.} for performing the sentence-level annotation. 
From the 36 articles, 4668 sentences are obtained. However, annotating all the sentences would be a huge task; hence we select a subset of  `important' sentences. 
In the subset first, all the sentences containing a negative sentiment word is added. Next, we use the sentence ranking component of the six existing extractive text summarization techniques. These techniques are categorized into: (i) graph-based~\cite{mihalcea2004textrank} (one method), (ii) feature-based~\cite{jagadeesh2005sentence} (two methods), and topic-based~\cite{ozsoylsa2011} (three methods), each outputs the top three `important' sentences.  The sentences which appear in the ranking list of any one of the algorithms is also added to the subset.  In this way, on an average 40\% of all sentences are shortlisted for annotation. 

The annotators are first subjected to an acclimatization task. They are provided with 
sentences and asked to independently 
identify the number of dimensions that mismatch with the reference brand personality (also provided to them). 
 When the annotators disagree on a particular sentence annotation, it is resolved through discussion until both of them come to a unanimous decision. {There was initial disagreement mostly when a sentence was ambiguous about its reference to the target brand - it was decided that such ambiguous sentences need to be seen in a larger context before marking them as inconsistent. 
Also, it was observed that some inconsistent sentences contained key informational content, without which the article is incomplete,
and hence should not be marked for modification.
}
Once their results begin to converge, 
then they are provided with the task of annotating the 36 articles which involve for each post identifying at least one sentence and at most three sentences that should be modified to {best} improve the brand consistency for the respective company keeping the key information content of the article intact. 
The final sentence set is the union of the sentences selected by the two annotators. 
We observe that the annotators agree for over 80\% of cases; in case of a mismatch and more than three sentences are present in the union set, an expert is referred to break the tie so that at most three sentences per article is selected as GOLD.

\subsection{Baselines}
We outline the baseline models.
\begin{enumerate}[(i)]
    \item RAND-3: We randomly select three sentences from the web article.
    \item LEAD-3: We select the leading (starting) three sentences because usually the important information is concentrated at the start of the article. It is a standard baseline model for a summarization task~\cite{nallapati2017summarunner}.
    \item CTR-3: We first select the sentences that have `whetherCentral' as positive and then sort it in descending of the number of `central' target entities contained in the sentence.
    \item POL-3: We first select the sentences that have `whetherNeg' as one and further sort them in descending order of `negScr' value.
\item CONS-3: We first order all the sentences in non-decreasing order of \textit{binLabelSim} and recommend the first three sentences of them all.
\item CONS-POL-3: We first rank in the same manner as CONS-3 and then perform a stable sorting based on `negScr' value. We first identify the sentences containing a negative sentiment and further rank them in decreasing order of negative score of a sentence.
\end{enumerate}

\subsection{Metrics}
For each web article, we now have a list of sentences which are manually selected by the human annotators (GOLD), and the sentences selected based on our methodology. To measure performance, we compute  n-gram overlap by (i) F1-ROUGE~\cite{lin2003automatic}: in order to assess informativeness; (ii) ROUGE-LCS: in order to assess the fluency by finding the longest common subsequence. 
We also evaluate the ability of the models to find the exact sentences to be recommended by  (iii) Precision [Prec@k] which
measures the fraction of top-k marked sentence (by an algorithm) to be present in GOLD. 
Prec@3 signify the precision score. 

\subsection{Experimental Results}

We compare the performance of our proposed model (MASR-3) with the baseline models in Table~\ref{tab:recco_comp}. 
We find that the ranking based scheme performs substantially better than LEAD-3 and RAND-3. 
MASR-3 significantly outperforms the single aspect-based ranking baselines except CTR (\textit{Centrality} aspect).
However, the Rouge-LCS score is high for all such schemes, as they are extractive in nature so fluency of the statements 
is naturally maintained. 
Also, the difference in Prec@1 score among different schemes is largely statistically insignificant. 
This implies that the topmost sentence returned by all the schemes has a similar rate of success in identifying one of the sentences from GOLD, proving the efficacy of each ranking scheme. 
The difference begins to appear as the value of $k$ is increased. 
Although MASR-3 is able to outperform CTR-3 across all metrics, the model improvement is not statistically significant. 
This can be partially 
 attributed to the small sample-size (limited set of 36 articles)  of the test dataset  whereby the significance test is passed when 
the performance difference is at least  0.1.
However, the better performance of CTR-3 proves the importance of centrality measure and justifies the design of sentence scoring function (Table \ref{tab:all3formula})
defined for MASR-3. 
The Prec@1 score is less for CTR-3, this happens in case of any additional measure, 
several 
 sentences have the same
relevance score and randomly tie-breaking is done to select the highest relevant sentence. This points  to the need for consideration of 
negative polarity and brand inconsistency. 

The overall performance of none of the schemes is not very high; on checking, we find a large part of this can be attributed to the inaccuracy in identifying the central sentences using the entity extraction technique. The scheme cannot extract any target entities or subjects for half of the sentences while in some cases, the co-reference resolution failed.
Thus even after correctly choosing a negative sentence, the miscalculation of the centrality score brings in error to the system.
The current tool does not comprehensively capture centrality; however, it re-establishes the proposition that centrality is a crucial aspect for the sentence ranking task.  
An immediate future work would be to develop a more sophisticated tool to identify the central sentences properly. 

\begin{table}[!ht]
\centering
\begin{tabular}{ c|c|c|c|c|c|c}
 \hline
  Models& ROUGE-1& ROUGE-2& ROUGE-LCS &$Prec@1$ &$Prec@2$&$Prec@3$ \\ \hline
RAND-3 &$0.395^{***}$&$0.251^{***}$&$0.762^{***}$&$0.167^{***}$&$0.222^{***}$&$0.204^{***}$\\
LEAD-3 &$0.433^{**}$&$0.297^{**}$&$0.771^{0.112}$&$0.472^{0.4}$&$0.306^{***}$&$0.278^{***}$\\ \hline
POL-3 &$0.492^{*}$&$0.372^{*}$&$0.789^{0.206}$&$0.5^{0.422}$&$0.472^{*}$&$0.389^{*}$\\
CONS-3 &$0.494^{**}$&$0.382^{*}$&$0.805^{0.206}$&$0.5^{0.487}$&$0.458^{*}$&$0.38^{*}$\\ 
CONS-POL-3 &$0.502^{*}$&$0.392^{0.083}$&$0.807^{0.276}$&$0.5^{0.487}$&$0.458^{*}$&$0.389^{*}$\\ 
CTR-3 &$0.54^{0.198}$&$0.435^{0.216}$&$0.821^{0.843}$&$0.472^{0.263}$& $0.542^{0.324}$&$0.468^{0.352}$\\
MASR-3 &\textbf{$0.571$}&\textbf{$0.472$}&\textbf{$0.823$}&\textbf{$0.556$}&\textbf{$0.583$}&\textbf{$0.5$}\\ \hline
\end{tabular}
\caption{Performance comparison for recommending sentences that need to be modified. We perform the paired t-test to determine whether the performance difference between MASR-3 and all the baseline model is statistically significant: *** indicates p-value $<0.001$, ** for p-value < 0.01, * for p-value between 0.01 and 0.05, and for p-values more than 0.05, we mention it directly in numerical form. For all models except CTR, the performance improvement by MASR-3 is statistically significant.}
\label{tab:recco_comp}
\end{table}

\section{Conclusion}\label{sec:conclusion}
To the best of our knowledge, this is the first work that characterizes the brand personality of an organization considering its corporate website. We collect data from 300K webpages covering around 650 companies, for a period of 17 years~(starting from January 2000) 
and then annotate a randomly chosen set of ~ 600 data points (both labelling and ranking the traits) using a crowdsourcing platform. We develop an independent classifier for each trait and use it to predict the branding information of a text automatically; the classification of the high confidence points is almost always correct.
With the classified dataset thus obtained, we study the brand characteristics of companies; we specifically study the roadblocks companies face in maintaining their brand image. 
We observe that some sectors, like media, struggle to maintain a consistent brand image. The brand image takes a beating, specifically when the initial branding is not very specific or topical; companies many time unwittingly damage their brand image by portraying a competent image during product promotion. Due to a company's strategy to extend the brand to multiple divergent products, the brand image suffers. We also find that it is immensely difficult to be consistent over time;  top-ranked Fortune 1000 companies are able to maintain it better but up to a certain span of time.
The interesting findings of the work thus can be used to flag content writers about the quality of consistency they are maintaining; we go one step further and propose a tool to recommend the top three sentences responsible for inconsistency if any.  The proposed model, which is computed based on centrality, negative polarity and brand consistency, outperforms all the baseline models.

\noindent \textit{Future work:} 
The future work would be oriented towards improving the machinery used for characterization of brand personality. 
One possible way of improving the accuracy of the simple individual brand-specific classification model is to improve further the linguistic feature set by incorporating the RepLab findings where tf-idf feature vector is extended and shows significant performance gains. Another way may be to learn all the traits jointly. It may be that one trait (weakly) implies one or more of the others, and thus joint-learning strategy may improve the classifier performance. Also, the present work is text-centric; an immediate future work would be to develop a multimodal classifier that considers website images along with the text. 
Since the domain of online reputation monitoring suffers from a lack of well-annotated data, we have to create a human-tagged data set painstakingly; it is difficult to take advantage of sophisticated deep learning techniques. 
A way out is to create high-quality silver tagged dataset;  
one  may use WordNet-Affect~\cite{strapparava2004wordnet}, SenticNet~\cite{cambria2018senticnet} or OntoSenticNet~\cite{ontosenticnet2018}, for developing domain-specific affective lexicons and then use it to weakly tag the documents. 
One can  use  embedding-based representations to further improve the classifier performance and 
develop better metrics based upon such embeddings. 
While developing the sentence ranking scheme (MASR-3), we realized that the `centrality' aspect is crucial to the sentence scoring function of MASR-3, thus exploring more sophisticated techniques to better identify the target entities, and improving the computation of the `centrality' aspect is important future work. The capacity of the current helper to work for longer articles (having more than 500 words) needs to be enhanced, and this can be done by increasing the length of individual instances from sentences to textual segments; may use paragraph separation or topic-based summarization models. Finally, the capability of the helper tool may be enhanced to make word or phrase-level suggestions and also to make it deployable. This would be an essential part of our future work.

\noindent \textit{Reproducibility:} We make the MT\textsubscript{large} dataset that includes the static and dynamic pages data (mentioned in Section 3.1) publicly available in Zenodo\footnote{\url{https://zenodo.org/record/3565079}}. We make all the data files and codes related with the `sentence ranking tool' publicly available in Github\footnote{\url{https://github.com/roysoumya/brand-consistency-framework}}.

\begin{acks}
We sincerely thank the anonymous reviewers for their valuable suggestions. Soumyadeep Roy is supported by the Institute PhD Fellowship at Indian Institute of Technology Kharagpur.
\end{acks}

\bibliographystyle{ACM-Reference-Format}
\bibliography{sample-acmsmall}


\begin{thebibliography}{41}


\ifx \showCODEN    \undefined \def \showCODEN     #1{\unskip}     \fi
\ifx \showDOI      \undefined \def \showDOI       #1{#1}\fi
\ifx \showISBNx    \undefined \def \showISBNx     #1{\unskip}     \fi
\ifx \showISBNxiii \undefined \def \showISBNxiii  #1{\unskip}     \fi
\ifx \showISSN     \undefined \def \showISSN      #1{\unskip}     \fi
\ifx \showLCCN     \undefined \def \showLCCN      #1{\unskip}     \fi
\ifx \shownote     \undefined \def \shownote      #1{#1}          \fi
\ifx \showarticletitle \undefined \def \showarticletitle #1{#1}   \fi
\ifx \showURL      \undefined \def \showURL       {\relax}        \fi
\providecommand\bibfield[2]{#2}
\providecommand\bibinfo[2]{#2}
\providecommand\natexlab[1]{#1}
\providecommand\showeprint[2][]{arXiv:#2}

\bibitem[\protect\citeauthoryear{Aaker}{Aaker}{1990}]%
        {aaker1990brand}
\bibfield{author}{\bibinfo{person}{David Aaker}.}
  \bibinfo{year}{1990}\natexlab{}.
\newblock \showarticletitle{Brand Extensions: The Good, the Bad, and the Ugly}.
\newblock \bibinfo{journal}{\emph{MIT Sloan Management Review}}
  \bibinfo{volume}{31}, \bibinfo{number}{4} (\bibinfo{date}{Summer}
  \bibinfo{year}{1990}), \bibinfo{pages}{47--56}.
\newblock
\showISBNx{0019848X}
\urldef\tempurl%
\url{https://search.proquest.com/docview/224963141?accountid=27562}
\showURL{%
\tempurl}


\bibitem[\protect\citeauthoryear{Aaker}{Aaker}{1997}]%
        {aaker1997}
\bibfield{author}{\bibinfo{person}{Jennifer~L. Aaker}.}
  \bibinfo{year}{1997}\natexlab{}.
\newblock \showarticletitle{Dimensions of brand personality}.
\newblock \bibinfo{journal}{\emph{Journal of Marketing Research}}
  \bibinfo{volume}{34}, \bibinfo{number}{3} (\bibinfo{date}{08}
  \bibinfo{year}{1997}), \bibinfo{pages}{347--356}.
\newblock
\showISBNx{00222437}
\urldef\tempurl%
\url{https://search.proquest.com/docview/235235096?accountid=27562}
\showURL{%
\tempurl}


\bibitem[\protect\citeauthoryear{Aggarwal}{Aggarwal}{2004}]%
        {aggarwal2004effects}
\bibfield{author}{\bibinfo{person}{Pankaj Aggarwal}.}
  \bibinfo{year}{2004}\natexlab{}.
\newblock \showarticletitle{The effects of brand relationship norms on consumer
  attitudes and behavior}.
\newblock \bibinfo{journal}{\emph{Journal of consumer research}}
  \bibinfo{volume}{31}, \bibinfo{number}{1} (\bibinfo{year}{2004}),
  \bibinfo{pages}{87--101}.
\newblock


\bibitem[\protect\citeauthoryear{Amig{\'o}, De~Albornoz, Chugur, Corujo,
  Gonzalo, Mart{\'\i}n, Meij, De~Rijke, and Spina}{Amig{\'o}
  et~al\mbox{.}}{2013}]%
        {amigo2013overview}
\bibfield{author}{\bibinfo{person}{Enrique Amig{\'o}},
  \bibinfo{person}{Jorge~Carrillo De~Albornoz}, \bibinfo{person}{Irina Chugur},
  \bibinfo{person}{Adolfo Corujo}, \bibinfo{person}{Julio Gonzalo},
  \bibinfo{person}{Tamara Mart{\'\i}n}, \bibinfo{person}{Edgar Meij},
  \bibinfo{person}{Maarten De~Rijke}, {and} \bibinfo{person}{Damiano Spina}.}
  \bibinfo{year}{2013}\natexlab{}.
\newblock \showarticletitle{Overview of replab 2013: Evaluating online
  reputation monitoring systems}. In \bibinfo{booktitle}{\emph{International
  conference of the cross-language evaluation forum for european languages}}.
  \bibinfo{pages}{333--352}.
\newblock
\urldef\tempurl%
\url{https://doi.org/10.1007/978-3-642-40802-1_31}
\showURL{%
\tempurl}


\bibitem[\protect\citeauthoryear{Cambria, Poria, Hazarika, and Kwok}{Cambria
  et~al\mbox{.}}{2018}]%
        {cambria2018senticnet}
\bibfield{author}{\bibinfo{person}{Erik Cambria}, \bibinfo{person}{Soujanya
  Poria}, \bibinfo{person}{Devamanyu Hazarika}, {and} \bibinfo{person}{Kenneth
  Kwok}.} \bibinfo{year}{2018}\natexlab{}.
\newblock \showarticletitle{SenticNet 5: Discovering conceptual primitives for
  sentiment analysis by means of context embeddings}. In
  \bibinfo{booktitle}{\emph{Thirty-Second AAAI Conference on Artificial
  Intelligence}}.
\newblock


\bibitem[\protect\citeauthoryear{Chawla, Bowyer, Hall, and Kegelmeyer}{Chawla
  et~al\mbox{.}}{2002}]%
        {Chawla2002}
\bibfield{author}{\bibinfo{person}{Nitesh~V. Chawla}, \bibinfo{person}{Kevin~W.
  Bowyer}, \bibinfo{person}{Lawrence~O. Hall}, {and} \bibinfo{person}{W.~Philip
  Kegelmeyer}.} \bibinfo{year}{2002}\natexlab{}.
\newblock \showarticletitle{SMOTE: Synthetic Minority Over-sampling Technique}.
\newblock \bibinfo{journal}{\emph{J. Artif. Int. Res.}} \bibinfo{volume}{16},
  \bibinfo{number}{1} (\bibinfo{date}{June} \bibinfo{year}{2002}),
  \bibinfo{pages}{321--357}.
\newblock
\showISSN{1076-9757}
\urldef\tempurl%
\url{http://dl.acm.org/citation.cfm?id=1622407.1622416}
\showURL{%
\tempurl}


\bibitem[\protect\citeauthoryear{Chen and Rodgers}{Chen and Rodgers}{2006}]%
        {Chen2006}
\bibfield{author}{\bibinfo{person}{Qimei Chen} {and} \bibinfo{person}{Shelly
  Rodgers}.} \bibinfo{year}{2006}\natexlab{}.
\newblock \showarticletitle{Development of an Instrument to Measure Web Site
  Personality}.
\newblock \bibinfo{journal}{\emph{Journal of Interactive Advertising}}
  \bibinfo{volume}{7}, \bibinfo{number}{1} (\bibinfo{year}{2006}),
  \bibinfo{pages}{4--46}.
\newblock
\urldef\tempurl%
\url{https://doi.org/10.1080/15252019.2006.10722124}
\showDOI{\tempurl}


\bibitem[\protect\citeauthoryear{Delin}{Delin}{2007}]%
        {delin2007}
\bibfield{author}{\bibinfo{person}{Judy Delin}.}
  \bibinfo{year}{2007}\natexlab{}.
\newblock \showarticletitle{Brand Tone of Voice}.
\newblock \bibinfo{journal}{\emph{Journal of Applied Linguistics}}
  \bibinfo{volume}{2}, \bibinfo{number}{1} (\bibinfo{year}{2007}).
\newblock
\showISSN{1743-1743}
\urldef\tempurl%
\url{https://journals.equinoxpub.com/index.php/JAL/article/view/1471}
\showURL{%
\tempurl}


\bibitem[\protect\citeauthoryear{Depecik, van Everdingen, and van
  Bruggen}{Depecik et~al\mbox{.}}{2014}]%
        {brandDivestment2014}
\bibfield{author}{\bibinfo{person}{Baris Depecik}, \bibinfo{person}{Yvonne~M.
  van Everdingen}, {and} \bibinfo{person}{Gerrit~H. van Bruggen}.}
  \bibinfo{year}{2014}\natexlab{}.
\newblock \showarticletitle{Firm Value Effects of Global, Regional, and Local
  Brand Divestments in Core and Non-Core Businesses}.
\newblock \bibinfo{journal}{\emph{Global Strategy Journal}}
  \bibinfo{volume}{4}, \bibinfo{number}{2} (\bibinfo{year}{2014}),
  \bibinfo{pages}{143--160}.
\newblock
\urldef\tempurl%
\url{https://doi.org/10.1111/j.2042-5805.2014.1074.x}
\showDOI{\tempurl}


\bibitem[\protect\citeauthoryear{Digman}{Digman}{1990}]%
        {digman1990}
\bibfield{author}{\bibinfo{person}{John~M Digman}.}
  \bibinfo{year}{1990}\natexlab{}.
\newblock \showarticletitle{Personality structure: Emergence of the five-factor
  model}.
\newblock \bibinfo{journal}{\emph{Annual review of psychology}}
  \bibinfo{volume}{41}, \bibinfo{number}{1} (\bibinfo{year}{1990}),
  \bibinfo{pages}{417--440}.
\newblock
\urldef\tempurl%
\url{https://www.annualreviews.org/doi/pdf/10.1146/annurev.ps.41.020190.002221}
\showURL{%
\tempurl}


\bibitem[\protect\citeauthoryear{Douglas, Mills, and Kavanaugh}{Douglas
  et~al\mbox{.}}{2007}]%
        {douglas2007exploring}
\bibfield{author}{\bibinfo{person}{Alecia~C. Douglas},
  \bibinfo{person}{Juline~E. Mills}, {and} \bibinfo{person}{Raphael
  Kavanaugh}.} \bibinfo{year}{2007}\natexlab{}.
\newblock \showarticletitle{Exploring the Use of Emotional Features at Romantic
  Destination Websites}. In \bibinfo{booktitle}{\emph{Information and
  Communication Technologies in Tourism 2007}},
  \bibfield{editor}{\bibinfo{person}{Marianna Sigala}, \bibinfo{person}{Luisa
  Mich}, {and} \bibinfo{person}{Jamie Murphy}} (Eds.).
  \bibinfo{publisher}{Springer Vienna}, \bibinfo{address}{Vienna},
  \bibinfo{pages}{331--340}.
\newblock
\showISBNx{978-3-211-69566-1}
\urldef\tempurl%
\url{https://doi.org/10.1007/978-3-211-69566-1_31}
\showURL{%
\tempurl}


\bibitem[\protect\citeauthoryear{Dragoni, Poria, and Cambria}{Dragoni
  et~al\mbox{.}}{2018}]%
        {ontosenticnet2018}
\bibfield{author}{\bibinfo{person}{Mauro Dragoni}, \bibinfo{person}{Soujanya
  Poria}, {and} \bibinfo{person}{Erik Cambria}.}
  \bibinfo{year}{2018}\natexlab{}.
\newblock \showarticletitle{OntoSenticNet: A Commonsense Ontology for Sentiment
  Analysis}.
\newblock \bibinfo{journal}{\emph{IEEE Intelligent Systems}}
  \bibinfo{volume}{33}, \bibinfo{number}{3} (\bibinfo{year}{2018}),
  \bibinfo{pages}{77--85}.
\newblock
\showISSN{1541-1672}
\urldef\tempurl%
\url{https://doi.org/10.1109/MIS.2018.033001419}
\showDOI{\tempurl}


\bibitem[\protect\citeauthoryear{Finkel, Grenager, and Manning}{Finkel
  et~al\mbox{.}}{2005}]%
        {finkel2005incorporating}
\bibfield{author}{\bibinfo{person}{Jenny~Rose Finkel}, \bibinfo{person}{Trond
  Grenager}, {and} \bibinfo{person}{Christopher Manning}.}
  \bibinfo{year}{2005}\natexlab{}.
\newblock \showarticletitle{Incorporating Non-local Information into
  Information Extraction Systems by Gibbs Sampling}. In
  \bibinfo{booktitle}{\emph{Proceedings of the 43rd Annual Meeting on
  Association for Computational Linguistics}} \emph{(\bibinfo{series}{ACL
  '05})}. \bibinfo{publisher}{Association for Computational Linguistics},
  \bibinfo{address}{Stroudsburg, PA, USA}, \bibinfo{pages}{363--370}.
\newblock
\urldef\tempurl%
\url{https://doi.org/10.3115/1219840.1219885}
\showDOI{\tempurl}


\bibitem[\protect\citeauthoryear{Hutto and Gilbert}{Hutto and Gilbert}{2014}]%
        {HuttoVader2014}
\bibfield{author}{\bibinfo{person}{Clayton~J. Hutto} {and}
  \bibinfo{person}{Eric Gilbert}.} \bibinfo{year}{2014}\natexlab{}.
\newblock \showarticletitle{{VADER:} {A} Parsimonious Rule-Based Model for
  Sentiment Analysis of Social Media Text}. In
  \bibinfo{booktitle}{\emph{Proceedings of the Eighth International Conference
  on Weblogs and Social Media, {ICWSM} 2014, Ann Arbor, Michigan, USA, June
  1-4, 2014.}}
\newblock
\urldef\tempurl%
\url{http://www.aaai.org/ocs/index.php/ICWSM/ICWSM14/paper/view/8109}
\showURL{%
\tempurl}


\bibitem[\protect\citeauthoryear{Jagadeesh, Pingali, and Varma}{Jagadeesh
  et~al\mbox{.}}{2005}]%
        {jagadeesh2005sentence}
\bibfield{author}{\bibinfo{person}{J Jagadeesh}, \bibinfo{person}{Prasad
  Pingali}, {and} \bibinfo{person}{Vasudeva Varma}.}
  \bibinfo{year}{2005}\natexlab{}.
\newblock \showarticletitle{Sentence extraction based single document
  summarization}.
\newblock \bibinfo{journal}{\emph{International Institute of Information
  Technology, Hyderabad, India}}  \bibinfo{volume}{5} (\bibinfo{year}{2005}).
\newblock
\urldef\tempurl%
\url{http://web2py.iiit.ac.in/publications/default/download/inproceedings.pdf.60ed1ced-3d36-43f0-b4d3-a1f48519166f.pdf}
\showURL{%
\tempurl}


\bibitem[\protect\citeauthoryear{Keller}{Keller}{1999}]%
        {Keller1999}
\bibfield{author}{\bibinfo{person}{Kevin~Lane Keller}.}
  \bibinfo{year}{1999}\natexlab{}.
\newblock \showarticletitle{Managing Brands for the Long Run: Brand
  Reinforcement and Revitalization Strategies}.
\newblock \bibinfo{journal}{\emph{California Management Review}}
  \bibinfo{volume}{41}, \bibinfo{number}{3} (\bibinfo{year}{1999}),
  \bibinfo{pages}{102--124}.
\newblock
\urldef\tempurl%
\url{https://doi.org/10.2307/41165999}
\showDOI{\tempurl}


\bibitem[\protect\citeauthoryear{Keller}{Keller}{2009}]%
        {Keller2009}
\bibfield{author}{\bibinfo{person}{Kevin~Lane Keller}.}
  \bibinfo{year}{2009}\natexlab{}.
\newblock \showarticletitle{Building strong brands in a modern marketing
  communications environment}.
\newblock \bibinfo{journal}{\emph{Journal of Marketing Communications}}
  \bibinfo{volume}{15}, \bibinfo{number}{2-3} (\bibinfo{year}{2009}),
  \bibinfo{pages}{139--155}.
\newblock
\urldef\tempurl%
\url{https://doi.org/10.1080/13527260902757530}
\showDOI{\tempurl}


\bibitem[\protect\citeauthoryear{Kincaid, Fishburne~Jr, Rogers, and
  Chissom}{Kincaid et~al\mbox{.}}{1975}]%
        {kincaid1975derivation}
\bibfield{author}{\bibinfo{person}{J~Peter Kincaid}, \bibinfo{person}{Robert~P
  Fishburne~Jr}, \bibinfo{person}{Richard~L Rogers}, {and}
  \bibinfo{person}{Brad~S Chissom}.} \bibinfo{year}{1975}\natexlab{}.
\newblock \showarticletitle{Derivation of new readability formulas (automated
  readability index, fog count and flesch reading ease formula) for navy
  enlisted personnel}.
\newblock  (\bibinfo{year}{1975}).
\newblock


\bibitem[\protect\citeauthoryear{Lahiri}{Lahiri}{2015}]%
        {lahiri2015squinky}
\bibfield{author}{\bibinfo{person}{Shibamouli Lahiri}.}
  \bibinfo{year}{2015}\natexlab{}.
\newblock \showarticletitle{SQUINKY! A Corpus of Sentence-level Formality,
  Informativeness, and Implicature}.
\newblock \bibinfo{journal}{\emph{arXiv preprint arXiv:1506.02306}}
  (\bibinfo{year}{2015}).
\newblock


\bibitem[\protect\citeauthoryear{Lin and Hovy}{Lin and Hovy}{2003}]%
        {lin2003automatic}
\bibfield{author}{\bibinfo{person}{Chin-Yew Lin} {and} \bibinfo{person}{Eduard
  Hovy}.} \bibinfo{year}{2003}\natexlab{}.
\newblock \showarticletitle{Automatic Evaluation of Summaries Using N-gram
  Co-occurrence Statistics}. In \bibinfo{booktitle}{\emph{Proceedings of the
  2003 Human Language Technology Conference of the North {A}merican Chapter of
  the Association for Computational Linguistics}}. \bibinfo{pages}{150--157}.
\newblock
\urldef\tempurl%
\url{https://www.aclweb.org/anthology/N03-1020}
\showURL{%
\tempurl}


\bibitem[\protect\citeauthoryear{Liu, Xu, Gou, Liu, Akkiraju, and Shen}{Liu
  et~al\mbox{.}}{2016}]%
        {Liu2016}
\bibfield{author}{\bibinfo{person}{X. Liu}, \bibinfo{person}{A. Xu},
  \bibinfo{person}{L. Gou}, \bibinfo{person}{H. Liu}, \bibinfo{person}{R.
  Akkiraju}, {and} \bibinfo{person}{H. Shen}.} \bibinfo{year}{2016}\natexlab{}.
\newblock \showarticletitle{SocialBrands: Visual analysis of public perceptions
  of brands on social media}. In \bibinfo{booktitle}{\emph{2016 IEEE Conference
  on Visual Analytics Science and Technology (VAST)}}. \bibinfo{pages}{71--80}.
\newblock
\urldef\tempurl%
\url{https://doi.org/10.1109/VAST.2016.7883513}
\showDOI{\tempurl}


\bibitem[\protect\citeauthoryear{Liu, Xu, Wang, Schoudt, Mahmud, and
  Akkiraju}{Liu et~al\mbox{.}}{2017}]%
        {Liu2017}
\bibfield{author}{\bibinfo{person}{Zhe Liu}, \bibinfo{person}{Anbang Xu},
  \bibinfo{person}{Yi Wang}, \bibinfo{person}{Jerald Schoudt},
  \bibinfo{person}{Jalal Mahmud}, {and} \bibinfo{person}{Rama Akkiraju}.}
  \bibinfo{year}{2017}\natexlab{}.
\newblock \showarticletitle{Does Personality Matter?: A Study of Personality
  and Situational Effects on Consumer Behavior}. In
  \bibinfo{booktitle}{\emph{Proceedings of the 28th ACM Conference on Hypertext
  and Social Media}} \emph{(\bibinfo{series}{HT '17})}.
  \bibinfo{pages}{185--193}.
\newblock
\showISBNx{978-1-4503-4708-2}
\urldef\tempurl%
\url{https://doi.org/10.1145/3078714.3078733}
\showDOI{\tempurl}


\bibitem[\protect\citeauthoryear{Mairesse, Walker, Mehl, and Moore}{Mairesse
  et~al\mbox{.}}{2007}]%
        {mairesse2007}
\bibfield{author}{\bibinfo{person}{Fran\c{c}ois Mairesse},
  \bibinfo{person}{Marilyn~A. Walker}, \bibinfo{person}{Matthias~R. Mehl},
  {and} \bibinfo{person}{Roger~K. Moore}.} \bibinfo{year}{2007}\natexlab{}.
\newblock \showarticletitle{Using Linguistic Cues for the Automatic Recognition
  of Personality in Conversation and Text}.
\newblock \bibinfo{journal}{\emph{J. Artif. Int. Res.}} \bibinfo{volume}{30},
  \bibinfo{number}{1} (\bibinfo{date}{Nov.} \bibinfo{year}{2007}),
  \bibinfo{pages}{457--500}.
\newblock
\showISSN{1076-9757}
\urldef\tempurl%
\url{http://dl.acm.org/citation.cfm?id=1622637.1622649}
\showURL{%
\tempurl}


\bibitem[\protect\citeauthoryear{Majumder, Poria, Gelbukh, and
  Cambria}{Majumder et~al\mbox{.}}{2017}]%
        {Majumder2017}
\bibfield{author}{\bibinfo{person}{N. Majumder}, \bibinfo{person}{S. Poria},
  \bibinfo{person}{A. Gelbukh}, {and} \bibinfo{person}{E. Cambria}.}
  \bibinfo{year}{2017}\natexlab{}.
\newblock \showarticletitle{Deep Learning-Based Document Modeling for
  Personality Detection from Text}.
\newblock \bibinfo{journal}{\emph{IEEE Intelligent Systems}}
  \bibinfo{volume}{32}, \bibinfo{number}{2} (\bibinfo{year}{2017}),
  \bibinfo{pages}{74--79}.
\newblock
\showISSN{1541-1672}
\urldef\tempurl%
\url{https://doi.org/10.1109/MIS.2017.23}
\showDOI{\tempurl}


\bibitem[\protect\citeauthoryear{Malmi, Krause, Rothe, Mirylenka, and
  Severyn}{Malmi et~al\mbox{.}}{2019}]%
        {malmi2019encode}
\bibfield{author}{\bibinfo{person}{Eric Malmi}, \bibinfo{person}{Sebastian
  Krause}, \bibinfo{person}{Sascha Rothe}, \bibinfo{person}{Daniil Mirylenka},
  {and} \bibinfo{person}{Aliaksei Severyn}.} \bibinfo{year}{2019}\natexlab{}.
\newblock \showarticletitle{Encode, Tag, Realize: High-Precision Text Editing}.
  In \bibinfo{booktitle}{\emph{Proceedings of the 2019 Conference on Empirical
  Methods in Natural Language Processing and the 9th International Joint
  Conference on Natural Language Processing (EMNLP-IJCNLP)}}.
  \bibinfo{publisher}{Association for Computational Linguistics},
  \bibinfo{address}{Hong Kong, China}, \bibinfo{pages}{5053--5064}.
\newblock
\urldef\tempurl%
\url{https://doi.org/10.18653/v1/D19-1510}
\showDOI{\tempurl}


\bibitem[\protect\citeauthoryear{Mazloom, Rietveld, Rudinac, Worring, and van
  Dolen}{Mazloom et~al\mbox{.}}{2016}]%
        {MazloomMultimodal2016}
\bibfield{author}{\bibinfo{person}{Masoud Mazloom}, \bibinfo{person}{Robert
  Rietveld}, \bibinfo{person}{Stevan Rudinac}, \bibinfo{person}{Marcel
  Worring}, {and} \bibinfo{person}{Willemijn van Dolen}.}
  \bibinfo{year}{2016}\natexlab{}.
\newblock \showarticletitle{Multimodal Popularity Prediction of Brand-related
  Social Media Posts}. In \bibinfo{booktitle}{\emph{Proceedings of the 24th ACM
  International Conference on Multimedia}} \emph{(\bibinfo{series}{MM '16})}.
  \bibinfo{pages}{197--201}.
\newblock
\showISBNx{978-1-4503-3603-1}
\urldef\tempurl%
\url{https://doi.org/10.1145/2964284.2967210}
\showDOI{\tempurl}


\bibitem[\protect\citeauthoryear{Mihalcea and Tarau}{Mihalcea and
  Tarau}{2004}]%
        {mihalcea2004textrank}
\bibfield{author}{\bibinfo{person}{Rada Mihalcea} {and} \bibinfo{person}{Paul
  Tarau}.} \bibinfo{year}{2004}\natexlab{}.
\newblock \showarticletitle{{T}ext{R}ank: Bringing Order into Text}. In
  \bibinfo{booktitle}{\emph{Proceedings of the 2004 Conference on Empirical
  Methods in Natural Language Processing}}. \bibinfo{publisher}{Association for
  Computational Linguistics}, \bibinfo{address}{Barcelona, Spain},
  \bibinfo{pages}{404--411}.
\newblock
\urldef\tempurl%
\url{https://www.aclweb.org/anthology/W04-3252}
\showURL{%
\tempurl}


\bibitem[\protect\citeauthoryear{M{\"u}ller and Chandon}{M{\"u}ller and
  Chandon}{2003}]%
        {muller2003impact}
\bibfield{author}{\bibinfo{person}{Brigitte M{\"u}ller} {and}
  \bibinfo{person}{Jean-Louis Chandon}.} \bibinfo{year}{2003}\natexlab{}.
\newblock \showarticletitle{The impact of visiting a brand website on brand
  personality}.
\newblock \bibinfo{journal}{\emph{Electronic Markets}} \bibinfo{volume}{13},
  \bibinfo{number}{3} (\bibinfo{year}{2003}), \bibinfo{pages}{210--221}.
\newblock
\urldef\tempurl%
\url{https://doi.org/10.1080/1019678032000108301}
\showDOI{\tempurl}


\bibitem[\protect\citeauthoryear{Munigala, Mishra, Tamilselvam, Khare,
  Dasgupta, and Sankaran}{Munigala et~al\mbox{.}}{2018}]%
        {munigala2018persuaide}
\bibfield{author}{\bibinfo{person}{Vitobha Munigala}, \bibinfo{person}{Abhijit
  Mishra}, \bibinfo{person}{Srikanth~G. Tamilselvam}, \bibinfo{person}{Shreya
  Khare}, \bibinfo{person}{Riddhiman Dasgupta}, {and} \bibinfo{person}{Anush
  Sankaran}.} \bibinfo{year}{2018}\natexlab{}.
\newblock \showarticletitle{PersuAIDE ! An Adaptive Persuasive Text Generation
  System for Fashion Domain}. In \bibinfo{booktitle}{\emph{Companion
  Proceedings of the The Web Conference 2018}} \emph{(\bibinfo{series}{WWW
  '18})}. \bibinfo{pages}{335--342}.
\newblock
\showISBNx{978-1-4503-5640-4}
\urldef\tempurl%
\url{https://doi.org/10.1145/3184558.3186345}
\showDOI{\tempurl}


\bibitem[\protect\citeauthoryear{Nallapati, Zhai, and Zhou}{Nallapati
  et~al\mbox{.}}{2017}]%
        {nallapati2017summarunner}
\bibfield{author}{\bibinfo{person}{Ramesh Nallapati}, \bibinfo{person}{Feifei
  Zhai}, {and} \bibinfo{person}{Bowen Zhou}.} \bibinfo{year}{2017}\natexlab{}.
\newblock \showarticletitle{SummaRuNNer: a recurrent neural network based
  sequence model for extractive summarization of documents}. In
  \bibinfo{booktitle}{\emph{Proceedings of the Thirty-First AAAI Conference on
  Artificial Intelligence}}. \bibinfo{pages}{3075--3081}.
\newblock
\urldef\tempurl%
\url{https://www.aaai.org/ocs/index.php/AAAI/AAAI17/paper/view/14636/14080}
\showURL{%
\tempurl}


\bibitem[\protect\citeauthoryear{Ozsoy, Alpaslan, and Cicekli}{Ozsoy
  et~al\mbox{.}}{2011}]%
        {ozsoylsa2011}
\bibfield{author}{\bibinfo{person}{Makbule~Gulcin Ozsoy},
  \bibinfo{person}{Ferda~Nur Alpaslan}, {and} \bibinfo{person}{Ilyas Cicekli}.}
  \bibinfo{year}{2011}\natexlab{}.
\newblock \showarticletitle{Text summarization using Latent Semantic Analysis}.
\newblock \bibinfo{journal}{\emph{J. Inf. Sci.}} \bibinfo{volume}{37},
  \bibinfo{number}{4} (\bibinfo{year}{2011}), \bibinfo{pages}{405--417}.
\newblock
\urldef\tempurl%
\url{https://doi.org/10.1177/0165551511408848}
\showDOI{\tempurl}


\bibitem[\protect\citeauthoryear{Pavlick and Tetreault}{Pavlick and
  Tetreault}{2016}]%
        {pavlick2016empirical}
\bibfield{author}{\bibinfo{person}{Ellie Pavlick} {and} \bibinfo{person}{Joel
  Tetreault}.} \bibinfo{year}{2016}\natexlab{}.
\newblock \showarticletitle{An Empirical Analysis of Formality in Online
  Communication}.
\newblock \bibinfo{journal}{\emph{Transactions of the Association for
  Computational Linguistics}}  \bibinfo{volume}{4} (\bibinfo{year}{2016}),
  \bibinfo{pages}{61--74}.
\newblock
\urldef\tempurl%
\url{https://doi.org/10.1162/tacl\_a\_00083}
\showDOI{\tempurl}


\bibitem[\protect\citeauthoryear{Roy, Ganguly, Sural, Chhaya, and
  Natarajan}{Roy et~al\mbox{.}}{2019}]%
        {Roy2019BrandPersona}
\bibfield{author}{\bibinfo{person}{Soumyadeep Roy}, \bibinfo{person}{Niloy
  Ganguly}, \bibinfo{person}{Shamik Sural}, \bibinfo{person}{Niyati Chhaya},
  {and} \bibinfo{person}{Anandhavelu Natarajan}.}
  \bibinfo{year}{2019}\natexlab{}.
\newblock \showarticletitle{Understanding Brand Consistency from Web Content}.
  In \bibinfo{booktitle}{\emph{Proceedings of the 10th ACM Conference on Web
  Science}} \emph{(\bibinfo{series}{WebSci '19})}. \bibinfo{pages}{245--253}.
\newblock
\showISBNx{978-1-4503-6202-3}
\urldef\tempurl%
\url{https://doi.org/10.1145/3292522.3326048}
\showDOI{\tempurl}


\bibitem[\protect\citeauthoryear{Schmitt}{Schmitt}{2012}]%
        {schmitt2012consumer}
\bibfield{author}{\bibinfo{person}{Bernd Schmitt}.}
  \bibinfo{year}{2012}\natexlab{}.
\newblock \showarticletitle{The consumer psychology of brands}.
\newblock \bibinfo{journal}{\emph{Journal of Consumer Psychology}}
  \bibinfo{volume}{22}, \bibinfo{number}{1} (\bibinfo{year}{2012}),
  \bibinfo{pages}{7 -- 17}.
\newblock
\showISSN{1057-7408}
\urldef\tempurl%
\url{https://doi.org/10.1016/j.jcps.2011.09.005}
\showDOI{\tempurl}


\bibitem[\protect\citeauthoryear{Still}{Still}{2018}]%
        {Still2018}
\bibfield{author}{\bibinfo{person}{Jeremiah~D. Still}.}
  \bibinfo{year}{2018}\natexlab{}.
\newblock \showarticletitle{Web page visual hierarchy: Examining Faraday's
  guidelines for entry points}.
\newblock \bibinfo{journal}{\emph{Computers in Human Behavior}}
  \bibinfo{volume}{84} (\bibinfo{year}{2018}), \bibinfo{pages}{352 -- 359}.
\newblock
\showISSN{0747-5632}
\urldef\tempurl%
\url{https://doi.org/10.1016/j.chb.2018.03.014}
\showDOI{\tempurl}


\bibitem[\protect\citeauthoryear{Strapparava, Valitutti,
  et~al\mbox{.}}{Strapparava et~al\mbox{.}}{2004}]%
        {strapparava2004wordnet}
\bibfield{author}{\bibinfo{person}{Carlo Strapparava},
  \bibinfo{person}{Alessandro Valitutti}, {et~al\mbox{.}}}
  \bibinfo{year}{2004}\natexlab{}.
\newblock \showarticletitle{Wordnet affect: an affective extension of
  wordnet.}. In \bibinfo{booktitle}{\emph{4th International Conference on
  Language Resources and Evaluation}}, Vol.~\bibinfo{volume}{4}.
  \bibinfo{pages}{40}.
\newblock
\urldef\tempurl%
\url{http://www.lrec-conf.org/proceedings/lrec2004/pdf/369.pdf}
\showURL{%
\tempurl}


\bibitem[\protect\citeauthoryear{Sun, Liu, Cao, Luo, and Shen}{Sun
  et~al\mbox{.}}{2018}]%
        {Sun2018}
\bibfield{author}{\bibinfo{person}{X. Sun}, \bibinfo{person}{B. Liu},
  \bibinfo{person}{J. Cao}, \bibinfo{person}{J. Luo}, {and} \bibinfo{person}{X.
  Shen}.} \bibinfo{year}{2018}\natexlab{}.
\newblock \showarticletitle{Who Am I? Personality Detection Based on Deep
  Learning for Texts}. In \bibinfo{booktitle}{\emph{2018 IEEE International
  Conference on Communications (ICC)}}. \bibinfo{pages}{1--6}.
\newblock
\showISSN{1938-1883}
\urldef\tempurl%
\url{https://doi.org/10.1109/ICC.2018.8422105}
\showDOI{\tempurl}


\bibitem[\protect\citeauthoryear{Tausczik and Pennebaker}{Tausczik and
  Pennebaker}{2010}]%
        {liwc2010}
\bibfield{author}{\bibinfo{person}{Yla~R Tausczik} {and}
  \bibinfo{person}{James~W Pennebaker}.} \bibinfo{year}{2010}\natexlab{}.
\newblock \showarticletitle{The psychological meaning of words: LIWC and
  computerized text analysis methods}.
\newblock \bibinfo{journal}{\emph{Journal of language and social psychology}}
  \bibinfo{volume}{29}, \bibinfo{number}{1} (\bibinfo{year}{2010}),
  \bibinfo{pages}{24--54}.
\newblock


\bibitem[\protect\citeauthoryear{Wu, Kim, Li, and Ma}{Wu et~al\mbox{.}}{2019}]%
        {brandpersonmobileui}
\bibfield{author}{\bibinfo{person}{Ziming Wu}, \bibinfo{person}{Taewook Kim},
  \bibinfo{person}{Quan Li}, {and} \bibinfo{person}{Xiaojuan Ma}.}
  \bibinfo{year}{2019}\natexlab{}.
\newblock \showarticletitle{Understanding and Modeling User-Perceived Brand
  Personality from Mobile Application UIs}. In
  \bibinfo{booktitle}{\emph{Proceedings of the 2019 {CHI} Conference on Human
  Factors in Computing Systems, {CHI} 2019}}. \bibinfo{pages}{213}.
\newblock
\urldef\tempurl%
\url{https://doi.org/10.1145/3290605.3300443}
\showDOI{\tempurl}


\bibitem[\protect\citeauthoryear{Xu and Bailey}{Xu and Bailey}{2012}]%
        {xu2012you}
\bibfield{author}{\bibinfo{person}{Anbang Xu} {and} \bibinfo{person}{Brian
  Bailey}.} \bibinfo{year}{2012}\natexlab{}.
\newblock \showarticletitle{What do you think?: a case study of benefit,
  expectation, and interaction in a large online critique community}. In
  \bibinfo{booktitle}{\emph{Proceedings of the ACM 2012 conference on Computer
  Supported Cooperative Work}}. \bibinfo{pages}{295--304}.
\newblock
\urldef\tempurl%
\url{https://doi.org/10.1145/2145204.2145252}
\showDOI{\tempurl}


\bibitem[\protect\citeauthoryear{Xu, Liu, Gou, Akkiraju, Mahmud, Sinha, Hu, and
  Qiao}{Xu et~al\mbox{.}}{2016}]%
        {Xu2016}
\bibfield{author}{\bibinfo{person}{Anbang Xu}, \bibinfo{person}{Haibin Liu},
  \bibinfo{person}{Liang Gou}, \bibinfo{person}{Rama Akkiraju},
  \bibinfo{person}{Jalal Mahmud}, \bibinfo{person}{Vibha Sinha},
  \bibinfo{person}{Yuheng Hu}, {and} \bibinfo{person}{Mu Qiao}.}
  \bibinfo{year}{2016}\natexlab{}.
\newblock \showarticletitle{Predicting Perceived Brand Personality with Social
  Media}. In \bibinfo{booktitle}{\emph{Proceedings of the Tenth International
  Conference on Web and Social Media}}. \bibinfo{pages}{436--445}.
\newblock
\urldef\tempurl%
\url{http://www.aaai.org/ocs/index.php/ICWSM/ICWSM16/paper/view/13078}
\showURL{%
\tempurl}


\end{thebibliography}


\end{document}